\documentclass[runningheads]{llncs}
\usepackage{graphicx}

\usepackage{tikz}
\usepackage{comment}
\usepackage{amsmath,amssymb} 
\usepackage{color}
\usepackage{orcidlink}

\usepackage{booktabs}
\usepackage{multirow}
\usepackage{tabularx}

\usepackage[accsupp]{axessibility}  

\begin{document}
\pagestyle{headings}
\mainmatter
\def\ECCVSubNumber{3200}  

\title{LoRD: Local 4D Implicit Representation for High-Fidelity Dynamic Human Modeling} 

\titlerunning{LoRD: Local 4D Representation for Dynamic Human}
%
%
\author{Boyan Jiang$^{1}$ \quad Xinlin Ren$^{1}$ \quad Mingsong Dou$^{2}$ \quad Xiangyang Xue$^{1 \dagger}$ \quad \\ Yanwei Fu$^{1 \dagger}$ \quad Yinda Zhang$^{2 \dagger}$ \\}

\institute{$^{1}$Fudan University \quad $^{2}$Google}

\authorrunning{B. Jiang et al.}
%

\maketitle

{\let\thefootnote\relax\footnotetext{ Boyan Jiang, Xinlin Ren and Xiangyang Xue are with School of Computer Science, Fudan University. Yanwei Fu is with 
School of Data Science, Fudan University.}}
{\let\thefootnote\relax\footnotetext{ $^{\dagger}$ Corresponding authors. E-mail: Yanwei Fu (yanweifu@fudan.edu.cn).}}

\begin{abstract}
    Recent progress in 4D implicit representation focuses on globally controlling the shape and motion with low dimensional latent vectors, which is prone to missing surface details and accumulating tracking error. While many deep local representations have shown promising results for 3D shape modeling, their 4D counterpart does not exist yet.
    In this paper, we fill this blank by proposing a novel \textbf{Lo}cal 4D implicit \textbf{R}epresentation for \textbf{D}ynamic clothed human, named \textbf{LoRD}, which has the merits of both 4D human modeling and local representation, and enables high-fidelity reconstruction with detailed surface deformations, such as clothing wrinkles. Particularly,
    our key insight is to encourage the network to learn the latent codes of local part-level representation, capable of explaining the local geometry and temporal deformations. 
    To make the inference at test-time, we first estimate the inner body skeleton motion to track local parts at each time step, and then optimize the latent codes for each part via auto-decoding based on different types of observed data. Extensive experiments demonstrate that the proposed method has strong capability for representing 4D human, and outperforms state-of-the-art methods on practical applications, including 4D reconstruction from sparse points, non-rigid depth fusion, both qualitatively and quantitatively. Please check out the project page for video and code: \href{https://boyanjiang.github.io/LoRD/}{https://boyanjiang.github.io/LoRD/}.
\end{abstract}

\section{Introduction}
Dynamic 3D human modeling has been a long-standing challenge to  3D vision and  graphics communities, as it  is critical to various applications, such as VR/AR, animation and robot simulation. Traditional methods leverage well-designed parametric model \cite{anguelov2005scape} and physics-based simulation \cite{selle2008robust,terzopoulos1987elastically,goldenthal2007efficient,gillette2015real} to model the inner human body and deformable outer cloth separately, but they typically demand huge engineering efforts and expensive computational cost. Recently, many learning based methods have been proposed \cite{lassner2017unite,guler2019holopose,mehta2018single,kanazawa2019learning,kocabas2020vibe,choi2020beyond,bhatnagar2020combining,saito2021scanimate}; unfortunately, some of these methods can not model fine-grained geometry details
beyond inner body, while the others only support frame-wise reconstruction to produce dynamic sequence.

\begin{figure}
	\centering \includegraphics[width=1.0\linewidth]{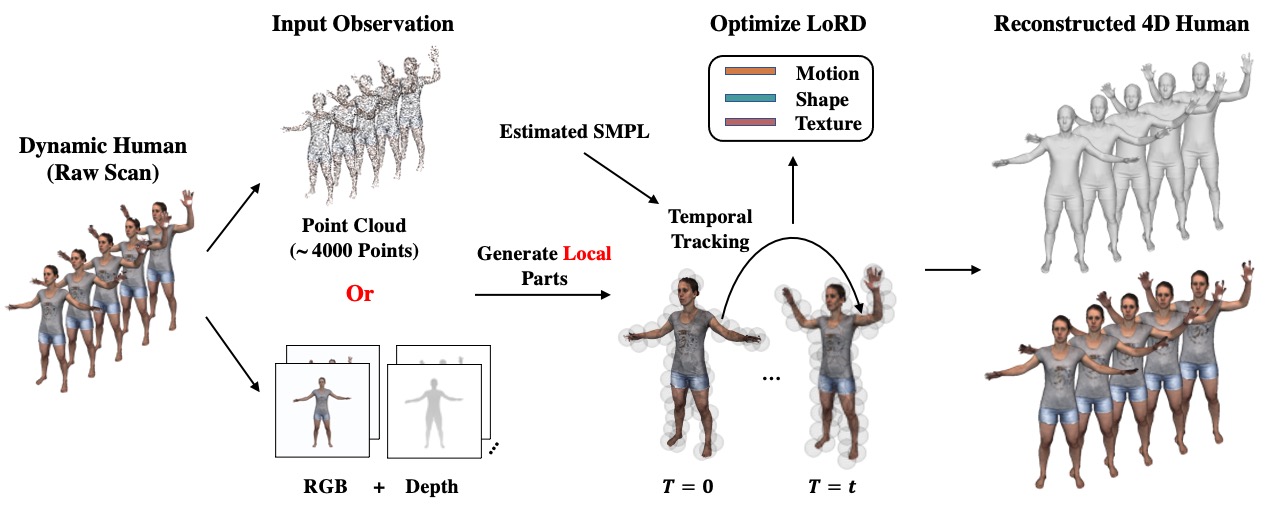}
	\caption{LoRD represents dynamic human with a set of overlapping local parts. Each part is temporally tracked with the estimated SMPL meshes, and contains low-dimensional latent codes of motion, canonical shape and texture (optional), which can be decoded to recover the detailed temporal changing of local surface patches by a 4D implicit network. During the test-time, these latent codes are optimized based on the different types of input observations, such as sparse point clouds and monocular RGB-D video to produce high-fidelity 4D human reconstruction.}
	\label{fig:teaser}
\end{figure}

The key challenge of dynamic human modeling is to find a way to model 4D representations for both surface geometry and temporal motion.
Typically, existing 4D human representation methods infer the single \textit{holistic} latent code/vector to control global motion and shape, which unfortunately are prone to over-smoothing shapes and missing fine-grained surface details. 
Recent efforts are made on
inferring local representations for 3D modeling~\cite{genova2020local,deng2020nasa,jiang2020local,chabra2020deep,peng2020convolutional}. Typically, these methods utilize a set of local parts to model the geometry of local surface regions for reconstructing complete 3D shapes.
Such local formulation improves the model capacity in recovering the detailed geometry with a stronger generalization ability than global free-form modeling \cite{niemeyer2019occupancy,mescheder2019occupancy,park2019deepsdf}.
However, it is nontrivial to directly enable these local methods to support the 4D scenario of modeling a dynamic 3D human with temporal motions, 
as their na\"ive extension to do per-frame reconstruction can not maintain the desirable properties of 4D modeling, such as temporal inter-/extrapolation, 4D spatial completion.

To this end, this paper proposes a \textbf{Lo}cal 4D implicit \textbf{R}epresentation for \textbf{D}ynamic human, named \textbf{LoRD}, which combines the merits of 4D human modeling and local representation. The LoRD is capable to produce high-fidelity human mesh sequence. Given a dynamic clothed human sequence over a time span $T \in \left[0,1\right]$, we decouple its temporal evolution into two factors: inner body skeleton motion and outer surface deformation. We handle the skeleton motion with the widely-used SMPL parametric model \cite{loper2015smpl}, which uses a shape parameter and a series of pose parameters to represent the temporal changing of inner body. On the other hand, for outer surface deformation, we resort to a local implicit framework. Specifically, we sample a bunch of local parts on the inner body mesh of the canonical frame ($T=0$), each part is represented by a 3D sphere with the intrinsic parameters (not camera intrinsics) of radius and transformation with respect to the world coordinate frame, and latent codes encoding local deformation and canonical shape information. 
Since SMPL models have the unified mesh topology, 
we can find the correspondence in subsequent frames and temporally align the local coordinate systems for each part.
Then we use a 4D local implicit network to model the surface deformation within each part conditioned on their latent codes. Such representation utilizes inner body model to handle the global skeleton motion, and leaves the detailed surface dynamics to the powerful local implicit network. This facilitates the dynamic human modeling with high-quality geometry.

Technically, our local representation is learned on 100 human sequences with ground truth mesh and its corresponding inner body mesh, each sequence contains $L=17$ frames. For each training sequence, we first sample the local parts on the surface of inner body mesh and randomly initialize the latent codes. Then we use objective function introduced by IGR \cite{gropp2020implicit} to optimize the local implicit network and latent codes. During the test-time, we fix the local implicit network to support a particular application (e.g., 4D reconstruction from sparse points, non-rigid depth fusion) via the auto-decoding method \cite{park2019deepsdf}. 
To obtain the inner body mesh, we use the existing work H4D \cite{jiang2022h4d} to provide plausible body estimation.
Moreover, our representation can combine with the H4D motion model to conduct body reference optimization introduced by PaMIR \cite{zheng2021pamir}, and support inner body refining to handle the imperfect body estimation (detailed in Sec. \ref{sec:optimization}). This improves the robustness of LoRD against inaccurate inner body tracking.

To summarize, the main contributions of our work are:
1) We propose a novel local 4D implicit representation, which divides surface of a dynamic human into a collection of local parts and supports high-fidelity dynamic human modeling;
2) To temporally align each part for training and test-time optimization, we leverage inner SMPL body mesh for local part tracking;
3) We design an inner body refining strategy based on our local representation to optimize imperfect initial body estimation;
4) Our representation only requires a small set of data for training, and outperforms the state-of-the-art methods on practical applications, e.g. 4D reconstruction from sparse points, non-rigid depth fusion.

\section{Related Work}
\noindent \textbf{4D representation}
Deep learning methods have shown impressive results on 3D-related tasks based on various representations, such as voxels \cite{choy20163d,GirdharFRG16,wang2017cnn}, point clouds \cite{qi2016pointnet,fan2017point,QiLWSG18,achlioptas2017representation}, meshes \cite{groueix2018atlasnet,kanazawa2018learning,pixel2mesh,liao2018deep,pixel2mesh++} and neural implicit surfaces \cite{mescheder2019occupancy,park2019deepsdf,chen2019learning,jiang2020local,chibane2020implicit,erler2020points2surf,chabra2020deep,genova2020local}. While great success has achieved for static 3D object, recent works \cite{niemeyer2019occupancy,rempe2020caspr,jiang2021learning} attempt to investigate elegant 4D representation of modeling dynamic 3D object with an additional temporal dimension. When targeting the dynamic human, recent methods \cite{niemeyer2019occupancy,jiang2021learning} always suffer from missing surface details and inaccurate motion due to the global shape modeling and lack of human motion prior. 
In contrast, the proposed local 4D representation leverages inner body tracking to handle the global skeleton motion and leaves the detailed dynamics to a set of local parts, which is effective to recover high-fidelity surface deformation, and generalize well to the novel sequences.

\noindent \textbf{Local shape representation}
The implicit representations conditioned on a global latent vector \cite{park2019deepsdf,mescheder2019occupancy} often produce over-smooth results and have failed to recover detailed geometry such as human hands and clothing wrinkles. To tackle this problem, some recent works utilize local implicit representation for shape modeling \cite{genova2020local,deng2020cvxnet,jiang2020local,peng2020convolutional} and neural rendering \cite{peng2021neural,lombardi2021mixture}, but none of them has used it to build 4D representation that represents how 3D geometry deforms \textit{continuously} over time. Similar to us, there is a family of work \cite{tiwari2021neural,chen2022gdna} building human avatar which supports shape generation under arbitrary body poses.
However, they process different timestamps independently and do not explicitly estimate temporal correspondences, which are shown to be important for recovering geometry details from multiple input frames or applications like motion completion/prediction.
In contrast, our method extends the local representation to 4D scenario by combining the human prior model and 4D implicit network, which can directly produce 4D results with one-shot optimization process.

\noindent \textbf{Dynamic human modeling}
When it comes to capturing the dynamic human, some methods \cite{xu2018monoperfcap,habermann2019livecap,habermann2020deepcap} require a pre-scanned template as a good initialization to obtain results from monocular color information. Recent methods \cite{newcombe2015dynamicfusion,yu2018doublefusion,zheng2018hybridfusion,su2020robustfusion} utilize depth sensors to achieve real-time speed based on the classical deformation graph \cite{sumner2007embedded} and volumetric fusion \cite{newcombe2011kinectfusion}, which get rid of subject-specific template. Since these methods are conducted in a frame-by-frame manner without intermediate motion representation, they are prone to error accumulation and hard to recover from tracking failures. Most recently, NDG \cite{bozic2021neural} learns a globally-consistent deformation graph to facilitate non-rigid reconstruction, but requires per-sequence retraining and relies on multi-view depth sensors, which is inconvenient in the actual usage. As a popular line of works, NeRF-based \cite{mildenhall2020nerf} human modeling methods \cite{peng2021neural,pumarola2021d} typically do not satisfy both local and temporal modeling. Most similar to us, Zheng et al. \cite{zheng2022structured} propose a structured temporal NeRF for dynamic human rendering. We note that these methods mainly focus on rendering quality but usually produce unsatisfactory geometry.
In contrast, LoRD models motion and shape jointly with local representation, so that information from two domains can be exchanged through the 4D model and benefit each other, which produces high-fidelity geometry results.

\begin{figure}
	\centering 
	\includegraphics[width=1.0\linewidth]{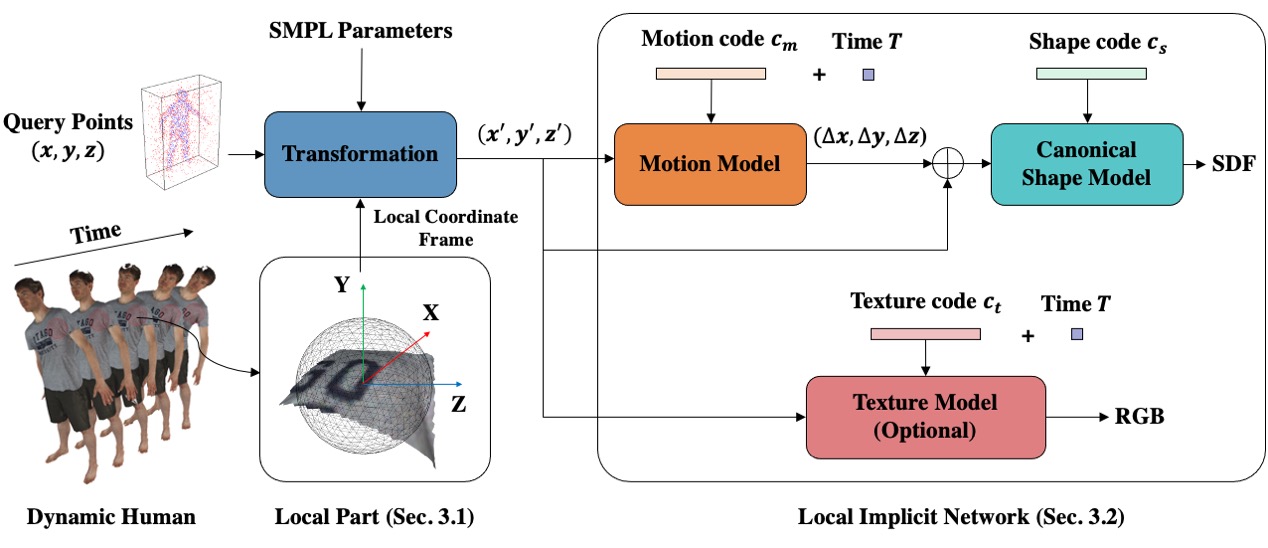}
	\caption{Overview of our framework. We use a set of spherical parts to model the local surface deformation of dynamic human. Given a 3D point $\left(x,y,z\right)$ under the world coordinate frame, we determine which part it falls into and transform it into the local coordinate frame, i.e. $\left(x',y',z'\right)$, according to the estimated SMPL parameters. The transformed point is queried into a local implicit network, which is conditioned on the latent codes of local part, to obtain signed distance and RGB (optional) value. Note that our local implicit network is shared by all parts. Meshes are extracted with Marching Cubes \cite{lorensen1987marching}.}
	\label{fig:pipeline}
\end{figure}

\section{Method}
 Our framework is overviewed in Fig.~\ref{fig:pipeline}: given a 3D clothed human mesh sequence of length $L=17$ frames that performs some motions in a normalized time span $\left[0,1\right]$, we first define a set of local parts (Sec. \ref{sec:local_part}) around inner body surface of the canonical frame ($T=0$ in our setup). Then we temporally track these parts which are controlled by the skeleton motion of the inner body model (SMPL). Note that we use the ground truth SMPL mesh during training, whereas the SMPL parameters are estimated with the off-the-shelf method \cite{jiang2022h4d} at test-time. Each part contains a motion code $c_m$, a shape code $c_s$ and a texture code $c_t$ (optional), which can be decoded by our local implicit network (Sec.~\ref{sec:local_net}) to obtain the reconstructed surface. Overall, we utilize the inner body model to track global skeleton motion and leave the detailed temporal deformation, geometry and texture of the local surface patch to the local implicit network. 
 Training and test-time optimization are discussed in Sec.~\ref{sec:training} and Sec.~\ref{sec:optimization}, respectively.

\subsection{Local Part Formulation} \label{sec:local_part}
\noindent \textbf{Inner body model} There are many ways to track the global skeleton motion of a dynamic human, e.g. optical/scene flow \cite{teed2020raft,liu2019flownet3d}, dense human correspondence \cite{wei2016dense,tan2021humangps}, and deformation graph \cite{sumner2007embedded}. In our formulation, we choose the widely-used SMPL model \cite{loper2015smpl} as it naturally provides surface correspondence between frames and its low-dimensional representations are easily to be optimized.

LoRD represents a 4D human with a set of local parts (defined as 3D spheres) $\mathcal{P}=\left\{\mathcal{P}_{k}\right\}_{k=1}^{K}$, where $\mathcal{P}_{k}=\left\{\mathbf{r},\mathbf{R}_k,\mathbf{c}_k\right\}$ is the intrinsic parameters of part $k$ (do not confuse them with camera intrinsics); $\mathbf{r} \in \mathbb{R}$ is the radius of the sphere shared by all parts (we use $r=5cm$ in our experiments); $\mathbf{R}_k \in \mathbb{R}^{9}$ and $\mathbf{c}_k \in \mathbb{R}^{3}$ are the rotation matrix relative to the world coordinate frame and the center of sphere for each part respectively. 
Given the inner body mesh of the canonical frame, a sampling algorithm (detailed in Supp.) is conducted on its surface to obtain the part centers. Inspired by \cite{jiang2020local,chabra2020deep}, to make the result smooth over the parts border, we use the overlapping strategy during the part sampling process, where each part overlaps with its neighboring parts by maximum $1.5\times$ the part radius $\mathbf{r}$, and finally produce 2127 parts.
The transformation of each part is based on the local coordinate frame as shown in Fig. \ref{fig:pipeline}. Details are in Supp. Mat.

\subsection{Local Implicit Network} \label{sec:local_net}
Besides the intrinsic parameters, each local part also has the latent parameters as low-dimensional codes $c_m$, $c_s$ and $c_t$, which encode respectively the information of the local surface deformation, canonical geometry and texture. The goal of the local parts is to represent the detailed temporal deformation and geometry of the local surface patches. To this end, we follow D-NeRF \cite{pumarola2021d} and use a 4D implicit network, which consists of a motion model and a canonical shape model. Moreover, if the observed data contain texture information, the additional texture model would be triggered to predict colors for the vertices of reconstructed mesh. Note that the implicit network is shared by all local parts. Next, We briefly introduce each model and the detailed architecture can be found in Supp. Mat.

\noindent \textbf{Motion model}
As shown in Fig. \ref{fig:pipeline}, we formulate the motion model $f^m\left(\mathbf{x},T\mid c_m\right)$ as a 4D function conditioned by the motion code $c_m \in \mathbb{R}^{128}$, which takes a 3D point $\mathbf{x}=\left(x,y,z\right)$ in the local coordinate frame and a time value $T$ (normalized to $\left[0,1\right]$) as input, and predicts a deformation vector $\mathrm{\Delta}\mathbf{x}$ that transforms this point to the canonical frame, i.e. $T=0$, by $\mathbf{x}^{\ast}=\mathbf{x}+\mathrm{\Delta}\mathbf{x}$. 
We adopt the network architecture of IM-Net \cite{chen2019learning}, and reduce the feature dimension of each hidden layer by 4 fold \cite{jiang2020local} to obtain an efficient motion model.

\noindent \textbf{Canonical shape model}
The canonical shape model $f^s\left(\mathbf{x}\mid c_s\right)$ is a neural signed distance function, which only holds a static implicit geometry of the canonical frame as the temporal deformation is handled by the motion model. Specifically, given a 3D query point at time $T$, we first obtain its position in the space of the canonical frame with the motion model, and then use the canonical shape model that is conditioned on a canonical shape latent code $c_s \in \mathbb{R}^{128}$ to predict the signed distance of the given point towards the surface. The same network architecture as DeepSDF \cite{park2019deepsdf} is adopted for canonical shape model. For training and testing efficiency, we reduce the number of layers and the feature channels for each layer to 6 and 256 respectively. During inference, we compute the bounding box of human based on the inner body mesh for each frame, and utilize the Marching Cubes algorithm \cite{lorensen1987marching} to extract the iso-surface.

\noindent \textbf{Texture model}
If the input data contains texture information, e.g. colored point clouds, our representation can be extended to support surface texture inference. We achieve this by learning a function $f^t\left(\mathbf{x}, T\mid c_t\right)$ to predict the 4D texture field \cite{oechsle2019texture,saito2019pifu,saito2021scanimate} of the dynamic local surface conditioned on a texture code $c_t  \in \mathbb{R}^{128}$. It takes a 3D point $\mathbf{x}$ in the local coordinate frame and a time value $T$, and outputs the RGB value of this point. We use the architecture of TextureField \cite{oechsle2019texture} decoder for our texture model. Please refer Supp. Mat. for the detailed network architecture. Note that we use our texture model in a per-sequence fashion during the test-time without pre-training, i.e. fit the input sequence with updating the network parameters, for better visualization results.

\subsection{Training} \label{sec:training}
Thank to our local formulation, the training of our model is very data efficient. We only use 100 sequences of length $L=17$ frames from CAPE dataset \cite{CAPE:CVPR:20} to learn our representation. During training, we adopt the auto-decoding method \cite{park2019deepsdf} and optimize our motion model, canonical shape model, and the latent codes for training parts. Specifically, given a training sequence that contains ground truth clothed meshes and the corresponding inner body meshes, we first sample a bunch of local parts on the surface of the inner body mesh of the first frame. Since the SMPL mesh has the unified surface topology, we can obtain the rotations and locations of each part in the following time steps, thus align their local coordinate frames. Next, we initialize the motion code and canonical shape code for each part with the vectors randomly sampled from $N\left(0,0.01\right)$, these codes are optimized with the network parameters during training. To train our implicit networks, the query points are sampled from three sources, i.e. surface, near surface space and free space in the bounding box. 

\noindent \textbf{Loss functions} The point sets sampled on-surface and off-surface are denoted as $\mathcal{X}$ and $\bar{\mathcal{X}}$ respectively. We optimize our 4D implicit function $f(\cdot)$ base on the loss functions introduced by IGR \cite{gropp2020implicit}: 
\begin{equation*}
    \mathcal{L}_{\mathrm{s}}=\frac{1}{|\mathcal{X}|}\sum_{\boldsymbol{x} \in \mathcal{X}}f(\boldsymbol{x})+\left.\left\|\nabla_{\boldsymbol{x}} f(\boldsymbol{x})-\boldsymbol{n}(\boldsymbol{x})\right\|\right), \quad \mathcal{L}_{\mathrm{e}}=\frac{1}{|\bar{\mathcal{X}}|}\sum_{\boldsymbol{x} \in \bar{\mathcal{X}}}\left(\left\|\nabla_{\boldsymbol{x}} f(\boldsymbol{x})\right\|-1\right)^{2}
\end{equation*}
where $\mathcal{L}_{\mathrm{s}}$ ensures the zero signed distance values for on-surface points and their normals aligned with the ground truth. $\mathcal{L}_{\mathrm{e}}$ is the regularization term encouraging the learned function to satisfy the Eikonal equation \cite{crandall1983viscosity}. In addition, we also add a latent regularization term $\mathcal{L}_{\mathrm{c}}=\left\|c_m\right\|_{2}+\left\|c_s\right\|_{2}$ to constrain the learning of latent spaces. The final objective function for training is $\mathcal{L}=\lambda_1\mathcal{L}_{\mathrm{s}}+\lambda_2\mathcal{L}_{\mathrm{e}}+\lambda_3\mathcal{L}_{\mathrm{c}}$. We use $\lambda_1=1.0$, $\lambda_2=1e^{-1}$, $\lambda_3=1e^{-3}$ in our experiment.

\noindent \textbf{Evaluate SDF for query points} During the training process, the sampled points are only evaluated by the local parts that cover them. In our case, ``point $\mathbf{x}$ is covered by part $k$'' means the Euclidean distance between $\mathbf{x}$ and the center of part $c_k$ is less than or equal to the pre-defined part radius $\mathbf{r}$, i.e. $d\left(\mathbf{x},c_k\right)\leq \mathbf{r}$. 
The sampled parts are highly overlapping, thus for one query point, we randomly choose $n$ parts that covered this point to evaluate its SDF, and then average $n$ SDF values ($n=4$ in our experiments) as the final output. This could encourage the network to produce the smooth results in the overlapping regions.
If some points are not covered by any parts, e.g. points sampled in the free space far from surface, then it will choose $n$-nearest parts to obtain the SDF prediction. Note that this is important for reconstructing complete results, since we cannot ensure the local parts sampled from inner body mesh would completely cover the surface of the clothed human.

\subsection{Test-Time Optimization} \label{sec:optimization}
After learning our local representation, we can then conduct the test-time optimization to reconstruct the dynamic human based on the given observations. In our experiments, we mainly focus on recovering 4D humans from complete point clouds or partial depth sequences. Generally speaking, the test-time optimization is similar to the training process, which performs backward optimization with the auto-decoding fashion, except that we fix the network parameters and only update the latent codes for each local part. Since we leverage the loss functions from IGR \cite{gropp2020implicit}, and directly perform optimization based on the point clouds with local-based representation, the geometry covered by each part is a non-watertight surface, which causes the extracted surface contains artificial interior back-faces. We borrow the post-processing algorithm from LIG \cite{jiang2020local} to remove such artifacts.
The details about the post-processing algorithm and the choices of hyper-parameters can be found in Supp. Mat. In addition, there are some technical details that we want to clarify below.

\noindent \textbf{Inner body estimation}
Given a testing sequence, we first need to estimate inner body meshes to sample local parts. As the temporal consistency could facilitate our reconstruction, we use the recent motion based human body estimation method H4D \cite{jiang2022h4d} to fit the SMPL parameters via backward optimization.

\noindent \textbf{Inner body refining}
The fitting results of H4D \cite{jiang2022h4d} are accurate enough in most cases, but still imperfect on some sequences, which may cause the observations of some local parts vary too much over time. Inspired by PaMIR~\cite{zheng2021pamir}, we propose a strategy to refine the initial inner body fitting from H4D. 
Specifically, we first sample and track the local parts on the initial body mesh sequence produced by H4D, and optimize the latent codes for each part. Then we fix the latent codes and local parts, query the SMPL vertices into our local implicit network, and optimize the SMPL parameters for shape and initial pose, and latent vector for motion of H4D. We follow the body reference optimization proposed in PaMIR to build the loss functions of our refining process:
\begin{equation*}
    \mathcal{L}_{\mathrm{SMPL}}=\left\{\begin{array}{ll}|f\left(x\right)| & f\left(x\right) \geq 0 \\ \frac{1}{\eta}|f\left(x\right)| & f\left(x\right)<0\end{array}\right., \quad \mathcal{L}_{reg}=\left\|V-V^{i n i t}\right\|_{2},
\end{equation*}
where $\eta=5$, $f\left(\cdot\right)$ is our local implicit signed distance function; $V=\left(\beta,\theta_0, c_m\right)$ contains the shape parameter, initial pose parameter and latent motion code of H4D, and the superscript ``init'' means initial estimations. This reflects the fact that, if the body estimation is accurate, then the vertices of the body mesh will get the negative SDF predictions (inside surface). Moreover, we also use an additional observation loss $\mathcal{L}_{\mathrm{obs}}$, which denotes Chamfer loss for the complete point cloud and point-to-surface loss for partial point cloud from the depth image. The final objective function is $\mathcal{L}=\lambda_1\mathcal{L}_{\mathrm{SMPL}}+\lambda_2\mathcal{L}_{\mathrm{obs}}+\lambda_3\mathcal{L}_{reg}$,
where $\lambda_1=1.0$, $\lambda_2=1e^{2}$ and $\lambda_3=1e^{-3}$ in our experiments. 
We verify the effectiveness of our inner body refining strategy in Sec. \ref{sec:ablation}.

\noindent \textbf{Texture model optimization}
As mentioned in Sec. \ref{sec:local_net}, we optimize the texture model for each testing sequence. Given a colored point cloud sequence, we can obtain the ground truth color $C_T\left(\mathbf{x}\right)$ of a surface point $\mathbf{x}$ in time $T$. Then we query $\mathbf{x}$ into the texture model conditioned on the texture code $c_t^k$ of part $k$ to get the color prediction. We also use the average of $n$ predicted colors as the final output (Sec. \ref{sec:training}). To optimize the network parameters and texture codes, we add the $L_1$-loss $\mathcal{L}_{\mathrm{color}}=\left|f^{c}(\mathbf{x},T\mid c_t)-C_T(\mathbf{x})\right|$ into the objective function.

\begin{figure}
	\centering
	\includegraphics[width=0.95\linewidth]{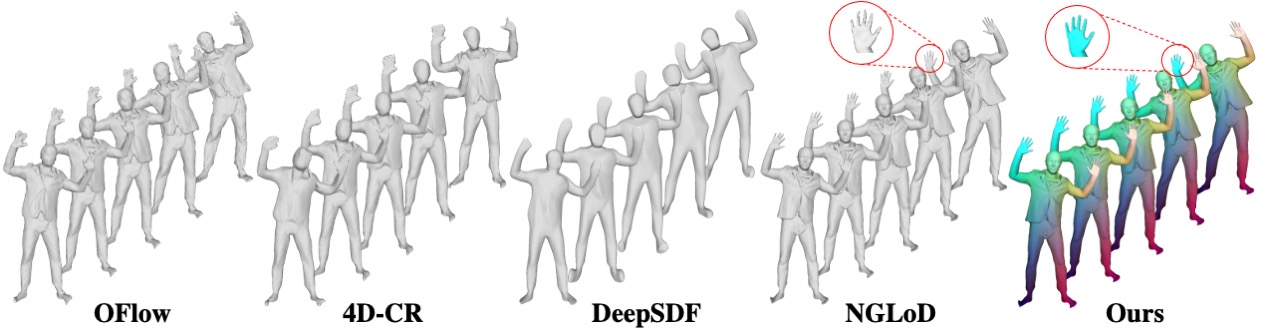}
	\caption{\label{fig:overfit}4D human fitting. We choose SoTA implicit 3D/4D representations to overfit a given mesh sequence and compare the results with us. The colors on our results indicate the correspondences across different frames, which cannot be obtained by the framewise baselines, i.e. NGLoD, DeepSDF. The zoomed-in part shows we reconstruct better finger details than NGLoD.}
\end{figure}

\section{Experiments}
In this section, we evaluate the representation capability of LoRD and its value in practical applications, i.e. 4D reconstruction and non-rigid depth fusion.

\begin{figure}
	\centering 
	\includegraphics[width=0.95\linewidth]{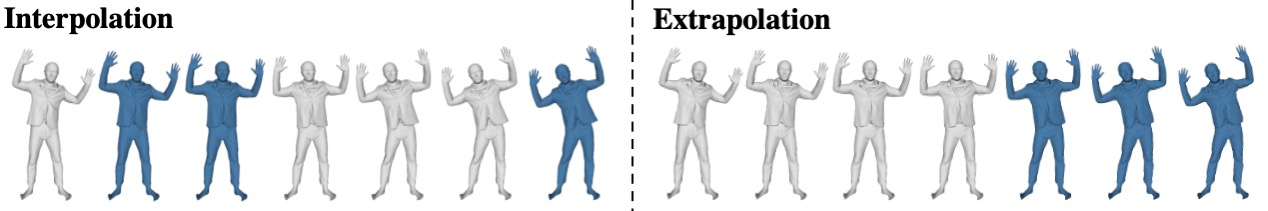}
	\caption{Temporal inter-/extrapolation. Colored meshes are inter-/extrapolated frames.}
	\label{fig:interpolation}
\end{figure}

\noindent \textbf{Dataset and metric}
For training and evaluation, we use the CAPE \cite{CAPE:CVPR:20} dataset which contains more than 600 motion sequences of 15 persons wearing different types of outfits, and the SMPL registrations are provided. Additionally, some raw scanned sequences with texture information are also available. We choose 100 sub-sequences of length $L=17$ for training, and use the sub-sequences of novel subjects for testing.
To compare with the baseline methods, we use Chamfer Distance-$L2$\cite{mescheder2019occupancy}, normal consistency \cite{saito2021scanimate} (the average $L2$ distance between the normal of given point on the source mesh and the normal of its nearest neighbor on the target mesh), and F-Score \cite{pixel2mesh} as evaluation metrics.

\begin{table}
\caption{\label{tab:overfit}Comparisons on 4D human fitting. 
Left: framewise methods, Right: temporal methods.
``Ch.-$L_2$'' and ``Normal'' mean Chamfer Distance ($\times 10^{-4}m^2$) and Surface Normal Consistency respectively. The threshold for computing F-Score is $\tau=5mm$.}
\begin{centering}
\begin{tabular}{cr}
{\scriptsize{}}%
\begin{tabular}{l}
{\scriptsize{}}%
\renewcommand\arraystretch{1}
\begin{tabular}{lccc}
\toprule[0.75px]
{\scriptsize{}Framewise} & {\scriptsize{}Ch.-$L_{2}$ $\downarrow$} & {\scriptsize{}Normal $\downarrow$} & {\scriptsize{}F-Score $\uparrow$}\tabularnewline
\midrule
\multirow{1}{*}[0.02\baselineskip]{{\scriptsize{}DeepSDF \cite{park2019deepsdf}}} & {\scriptsize{}0.846} & {\scriptsize{}0.291} & {\scriptsize{}0.669}\tabularnewline
{\scriptsize{}NGLoD \cite{takikawa2021neural}} & \textbf{\scriptsize{}0.074} & {\scriptsize{}0.135} & \textbf{\scriptsize{}0.969}\tabularnewline
\toprule[0.75px] 
\end{tabular}\tabularnewline
\end{tabular} & {\scriptsize{}}%
\renewcommand\arraystretch{0.7}
\begin{tabular}{r}
{\scriptsize{}}%
\begin{tabular}{lccc}
\toprule[0.75px] 
{\scriptsize{}Temporal} & {\scriptsize{}Ch.-$L_{2}$ $\downarrow$} & {\scriptsize{}Normal $\downarrow$} & {\scriptsize{}F-Score $\uparrow$}\tabularnewline
\midrule 
{\scriptsize{}OFlow \cite{niemeyer2019occupancy}} & {\scriptsize{}0.317} & {\scriptsize{}0.312} & {\scriptsize{}0.675}\tabularnewline
{\scriptsize{}4D-CR \cite{jiang2021learning}} & {\scriptsize{}5.249} & {\scriptsize{}0.359} & {\scriptsize{}0.425}\tabularnewline
{\scriptsize{}Ours} & {\scriptsize{}0.075} & \textbf{\scriptsize{}0.131} & \textbf{\scriptsize{}0.969}\tabularnewline
\toprule[0.75px] 
\end{tabular}\tabularnewline
\end{tabular}\tabularnewline
\end{tabular}
\par\end{centering}
\end{table}

\noindent \textbf{Implementation details} 
We use PyTorch with Adam optimizer \cite{kingma2014adam} of learning rate $1e^{-3}$ and batch size 1 for both training and test-time optimization. The experiments are conducted on a single Nvidia 2080Ti GPU. The test-time optimization takes around $15min$ for each 17 frames sequence.

\subsection{Representation Capability}
\noindent \textbf{4D human fitting}
We first evaluate the efficacy of LoRD in representing dynamic human by overfitting a given mesh sequence. We select one sequence from the CAPE dataset for this task. For comparison, we choose 3D neural SDF methods DeepSDF \cite{park2019deepsdf} and NGLoD \cite{takikawa2021neural}, DeepSDF is a global representation which represents the complete shape with a single latent code, while NGLoD is a SoTA local neural SDF representation based on the Octree, both of them are 3D representations that need to work with frame-wise manner to produce a temporal sequence. In addition, we choose the SoTA 4D representation methods OFlow \cite{niemeyer2019occupancy} and 4D-CR \cite{jiang2021learning} as our baseline.

The quantitative results are shown in Tab. \ref{tab:overfit}. Our LoRD representation clearly outperforms DeepSDF and all the SoTA 4D representation methods, and performs comparable with framewise method NGLoD. We show the visual results in Fig. \ref{fig:overfit}, the colors of our results indicate the dense correspondences w.r.t the first frame. Specifically, for each vertex on the reconstructed mesh of time $T$, we use the optimized motion codes to transform it to the first frame, and obtain color value of the nearest vertex. We note that this cannot be achieved by DeepSDF or NGLoD, since they do not model temporal information.

\noindent \textbf{Temporal inter-/extrapolation}
To further show the superiority of LoRD over the framewise representations, we show the temporal inter-/extrapolation results achieved by our method in Fig. \ref{fig:interpolation}. Given a sequence of length $L=17$ frames, for interpolation, we randomly choose 9 frames as the observations to perform SDF fitting, the goal is to complete the missing frames to obtain a temporally complete sequence. And for extrapolation, we only use the first 9 frames and need predict the future motion of the last 8 frames. Fig. \ref{fig:interpolation} shows that LoRD produces the plausible results on both inter- or extra-polation modes. Again, these temporal completion tasks also cannot be achieved by the framewise 3D representations, e.g. DeepSDF, NGLoD. We also provide the results about interpolation of the latent codes in Supp. Mat. (Sec. 2.2) as a sanity check.

\subsection{4D Reconstruction from Sparse Points} \label{sec:4d_recons}
We then show that LoRD can support various applications. First, we demonstrate that LoRD can achieve high quality 4D reconstruction from sparse point clouds. In this case, we assume the point normal directions are available (oriented point cloud, the same for Poisson Reconstruction \cite{kazhdan2013screened} and LIG \cite{jiang2020local}).

\begin{figure}
	\centering 
	\includegraphics[width=1.0\linewidth]{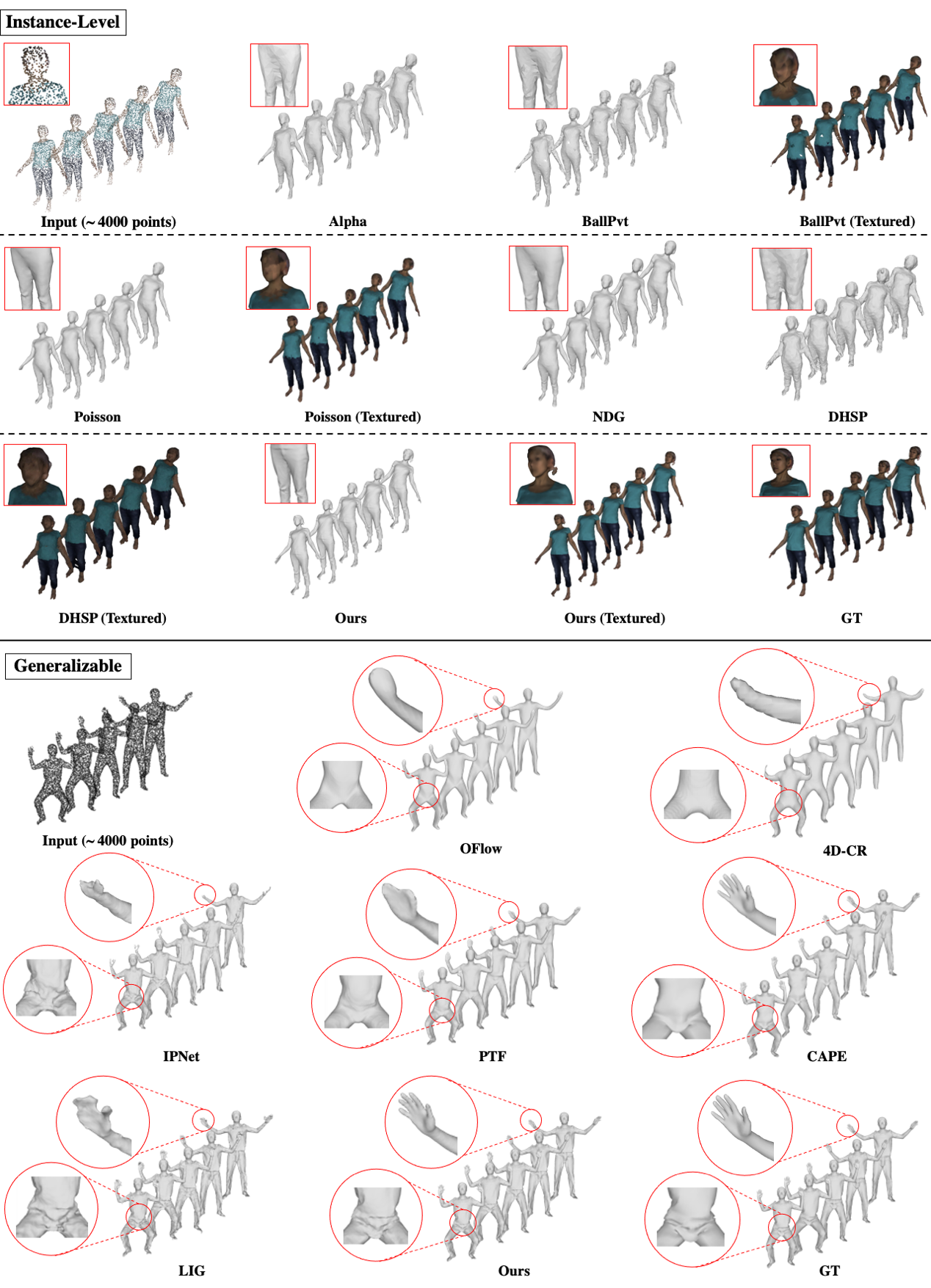}
	\caption{4D reconstruction from sparse points. Each input point cloud contains around 4000 points. Note the detailed geometry in the zoomed-in parts and the surface deformation recovered by our method. We provide more qualitative results in Supp. Mat.}
	\label{fig:4d_recons}
\end{figure}

\begin{figure}
\centering{}%
\begin{tabular}{cc}
\begin{tabular}{c}
{{}}%
\renewcommand\arraystretch{0.7}
\begin{tabular}{c|cccc}
\hline 
\multicolumn{5}{c}{\textit{\tiny{}I. Comparisons to instance-level methods}}\tabularnewline
\hline 
\multicolumn{1}{c}{{\tiny{}P./$m^{2}$}} & {\tiny{}Method} & {\tiny{}Ch.-$L_{2}$ $\downarrow$} & {\tiny{}Normal $\downarrow$} & {\tiny{}F-Score $\uparrow$}\tabularnewline
\hline 
\multirow{6}{*}{{\tiny{}500}} & {\tiny{}Alpha\cite{edelsbrunner1994three}} & {\tiny{}1.665} & {\tiny{}1.205} & {\tiny{}0.422}\tabularnewline
 & {\tiny{}BallPvt\cite{bernardini1999ball}} & {\tiny{}0.740} & {\tiny{}0.433} & {\tiny{}0.590}\tabularnewline
 & {\tiny{}PSR10 \cite{kazhdan2013screened}} & {\tiny{}0.664} & {\tiny{}0.310} & {\tiny{}0.714}\tabularnewline
 & {\tiny{}DHSP \cite{wei2021deep}} & {\tiny{}1.383} & {\tiny{}0.864} & {\tiny{}0.520}\tabularnewline
 & {\tiny{}NDG \cite{bozic2021neural}} & {\tiny{}0.706} & {\tiny{}0.2901} & {\tiny{}0.712}\tabularnewline
 & {\tiny{}Ours} & \textbf{\tiny{}0.105} & \textbf{\tiny{}0.176} & \textbf{\tiny{}0.938}\tabularnewline
\hline 
\multirow{6}{*}{{\tiny{}1000}} & {\tiny{}Alpha \cite{edelsbrunner1994three}} & {\tiny{}0.966} & {\tiny{}1.191} & {\tiny{}0.546}\tabularnewline
 & {\tiny{}BallPvt \cite{bernardini1999ball}} & {\tiny{}0.337} & {\tiny{}0.545} & {\tiny{}0.746}\tabularnewline
 & {\tiny{}PSR10 \cite{kazhdan2013screened}} & {\tiny{}0.301} & {\tiny{}0.271} & {\tiny{}0.822}\tabularnewline
 & {\tiny{}DHSP \cite{wei2021deep}} & {\tiny{}0.352} & {\tiny{}1.131} & {\tiny{}0.686}\tabularnewline
 & {\tiny{}NDG \cite{bozic2021neural}} & {\tiny{}0.316} & {\tiny{}0.254} & {\tiny{}0.819}\tabularnewline
 & {\tiny{}Ours} & \textbf{\tiny{}0.105} & \textbf{\tiny{}0.160} & \textbf{\tiny{}0.946}\tabularnewline
\hline 
\multirow{6}{*}{{\tiny{}2000}} & {\tiny{}Alpha \cite{edelsbrunner1994three}} & {\tiny{}0.343} & {\tiny{}1.160} & {\tiny{}0.726}\tabularnewline
 & {\tiny{}BallPvt \cite{bernardini1999ball}} & {\tiny{}0.187} & {\tiny{}0.546} & {\tiny{}0.860}\tabularnewline
 & {\tiny{}PSR10 \cite{kazhdan2013screened}} & {\tiny{}0.175} & {\tiny{}0.223} & {\tiny{}0.905}\tabularnewline
 & {\tiny{}DHSP \cite{wei2021deep}} & {\tiny{}0.181} & {\tiny{}0.607} & {\tiny{}0.808}\tabularnewline
 & {\tiny{}NDG \cite{bozic2021neural}} & {\tiny{}0.177} & {\tiny{}0.217} & {\tiny{}0.901}\tabularnewline
 & {\tiny{}Ours} & \textbf{\tiny{}0.102} & \textbf{\tiny{}0.154} & \textbf{\tiny{}0.952}\tabularnewline
\hline 
\hline 
\multicolumn{5}{c}{\textit{\tiny{}II. Comparisons to generalizable methods}}\tabularnewline
\hline 
\multicolumn{1}{c}{{\tiny{}Type}} & {\tiny{}Method} & {\tiny{}Ch.-$L_{2}$ $\downarrow$} & {\tiny{}Normal $\downarrow$} & {\tiny{}F-Score $\uparrow$}\tabularnewline
\hline 
\multirow{4}{*}{{\tiny{}Framewise}} & {\tiny{}IPNet \cite{bhatnagar2020combining}} & {\tiny{}0.752} & {\tiny{}0.298} & {\tiny{}0.572}\tabularnewline
 & {\tiny{}PTF \cite{wang2021locally}} & {\tiny{}0.582} & {\tiny{}0.278} & {\tiny{}0.485}\tabularnewline
 & {\tiny{}CAPE \cite{CAPE:CVPR:20}} & {\tiny{}0.749} & {\tiny{}0.332} & {\tiny{}0.411}\tabularnewline
 & {\tiny{}LIG \cite{jiang2020local}} & {\tiny{}0.623} & {\tiny{}0.289} & {\tiny{}0.875}\tabularnewline
\hline 
\multirow{3}{*}{{\tiny{}Temporal}} & {\tiny{}OFlow \cite{niemeyer2019occupancy}} & {\tiny{}5.767} & {\tiny{}0.344} & {\tiny{}0.350}\tabularnewline
 & {\tiny{}4D-CR \cite{jiang2021learning}} & {\tiny{}5.162} & {\tiny{}0.398} & {\tiny{}0.390}\tabularnewline
 & {\tiny{}Ours} & \textbf{\tiny{}0.306} & \textbf{\tiny{}0.204} & \textbf{\tiny{}0.908}\tabularnewline
\hline 
\end{tabular}\tabularnewline
\end{tabular} & {{}}%
\begin{tabular}{c}
\includegraphics[width=0.48\linewidth]{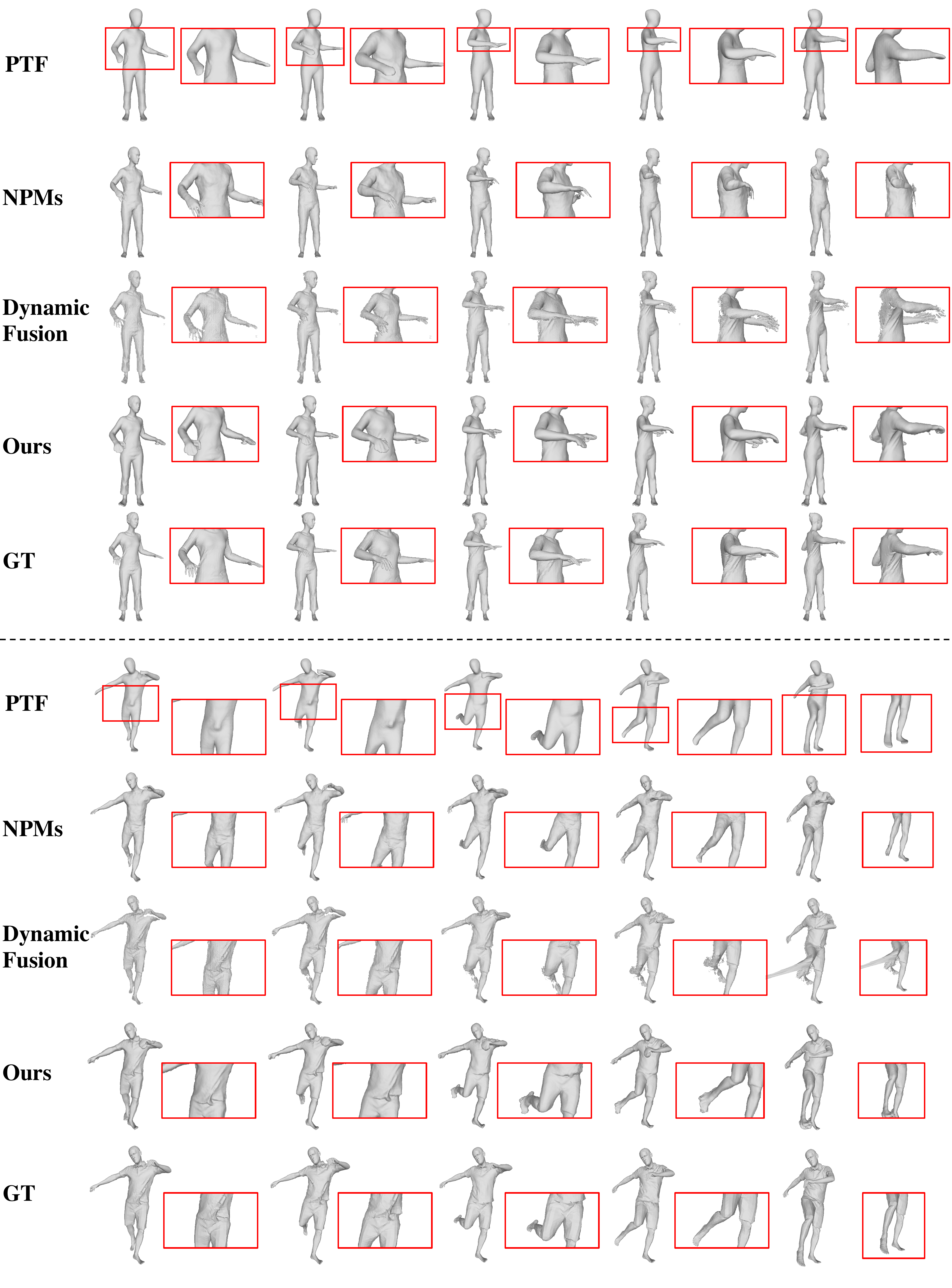}\tabularnewline
\end{tabular}\tabularnewline
(a) 4D reconstruction & (b) Non-rigid depth fusion \tabularnewline
\end{tabular}
\caption{ \label{fig:merged_fig}(a) Comparisons on 4D reconstruction from sparse points. The leftmost column in Block I represents the sampled point cloud density, the smaller number corresponds to the sparser point cloud. The results in Block II are obtained from the point cloud of density 2000 points/$m^2$. (b) Qualitative comparisons on non-rigid depth fusion.}
\end{figure}

\noindent \textbf{Compare to instance-level methods} We first compare LoRD with the instance level methods, the ``instance-level'' in here means we only overfit one sequence at a time and do not consider generalization to other instances. 
We choose the traditional Poisson Surface Reconstruction with octree depth value $d=10$ (PSR10) \cite{kazhdan2013screened}, Alpha Shape \cite{edelsbrunner1994three} and Ball Pivoting \cite{bernardini1999ball} as the baseline. Moreover, we also compare with the SoTA network-based surface reconstruction method Deep Hybrid Self-Prior (DHSP), and the non-rigid reconstruction method Neural Deformation Graph (NDG). The quantitative results are show in Fig. \ref{fig:merged_fig} (a, I), the leftmost column represents the sampled point cloud density (number of points per square meter of surface), the smaller number corresponds to the sparser point cloud, the surface area of SMPL mesh used for point sampling is around 2$m^2$. As can be seen, our method outperforms all the baselines by a large margin. More importantly, the sparser point cloud hardly affects our performance while the baseline methods have been significantly affected, this is because LoRD is a 4D representation, sparse observation from each frame can compensate each other through the motion model. The qualitative comparisons are shown in Fig. \ref{fig:4d_recons} (above the solid line), our method can recover geometry details on the face and cloth with high resolution texture, while the baselines only produce over-smooth results due to the limited information from sparse inputs.

\noindent \textbf{Compare to generalizable methods}
To show the generalization ability of our method, we train LoRD on the training set of 100 sequences, then fix the network parameters and optimize the latent codes of local parts to fit the input point cloud via back-propagation. In this experiment, we use the point density of 2000 points/$m^2$ (same as the results in the last group of Fig. \ref{fig:merged_fig} (a, I)), and choose 10 testing sequences of novel subjects for evaluation. As framewise baselines, we choose: IPNet \cite{bhatnagar2020combining} and PTF \cite{wang2021locally}, which takes point cloud as input and output reconstructed mesh via feed forward fashion; CAPE \cite{CAPE:CVPR:20} and LIG \cite{jiang2020local}, which obtain reconstructions via the backward optimization similar to us. The OFlow and 4D-CR are still considered as the baseline of temporal methods, we remove their encoders, fix the decoder parameters, and perform backward optimization. For OFlow and 4D-CR, we use the ground truth occupancy instead of oriented point cloud as supervision for more stable results.
The results are shown in Fig. \ref{fig:4d_recons} (below the
solid line) and Fig. \ref{fig:merged_fig} (a, II), our method beats all the baselines both qualitatively and quantitatively. We can observe the fine-grained geometry recovered by LoRD in the zoomed-in parts of Fig. \ref{fig:4d_recons}, as well as detailed clothing deformation, which show that our model trained on small set of data can generalize well to the novel motion sequences. More results are in Supp. Mat.

\subsection{Non-Rigid Depth Fusion}
We further test LoRD with the application of non-rigid depth fusion. Given a static RGB-D camera, with a person standing in front of it performing different actions, the goal is to accurately track the human motion and merge all depth observations in a time span, and finally produce a dynamic mesh sequence.
In this experiment, we use the mesh sequences of length $L=17$ from CAPE dataset \cite{CAPE:CVPR:20}, and render each frame to get depth image of resolution $512\times512$. We compute the normal map based on the depth image, and back-project each pixel into 3D space with the known camera intrinsics to obtain the partial oriented point cloud as the observations. Then we run H4D \cite{jiang2022h4d} to get the inner body estimation, and use our pretrained LoRD model to perform auto-decoding. Our approach formulates non-rigid fusion as a temporal completion problem within local parts.
We choose DynamicFusion \cite{newcombe2015dynamicfusion}, NPMs \cite{palafox2021npms} and PTF \cite{wang2021locally} as our baseline and show the qualitative comparisons in Fig. \ref{fig:merged_fig} (b).
We observe that PTF produces overly smooth results, NPMs cannot model the detailed surface geometry for different subjects, and DynamicFusion fails to track the human motion that is very fast or contains self-occlusion and leads to unsatisfactory fusions.
In contrast, our model is capable to produce more complete fusion results than DynamicFusion, e.g. back of the first example, and more detailed geometry than PTF and NPMs.
Additional results including non-rigid fusion on real-world data and the comparison to more recent human specific fusion work DoubleFusion \cite{yu2018doublefusion} are provided in Supp. Mat. for the sake of space. Our method shows robustness to the SMPL fitting error and provides more complete results than DoubleFusion.

\subsection{Ablation Study} \label{sec:ablation}
\noindent \textbf{Imperfect body tracking}
We first provide an ablation study to demonstrate the effectiveness of the proposed inner body refining method. We use the 4D reconstruction task with the point density 2000 point/$m^2$ for evaluation. Given the initially estimated SMPL inner body, we manually add the random Gaussian noise to it and compare the reconstruction performances before and after refining. Specifically, we perturb the SMPL shape ($\beta$) and pose ($\theta$) parameters by $\beta+=\lambda_{\beta} \cdot \sigma \cdot \mu$ and $\theta+=\lambda_{\theta} \cdot \sigma \cdot \mu$,
where $\mu \in N\left(0,1\right)$, $\lambda_{\beta}=0.05$, $\lambda_{\theta}=0.01$, and $\sigma \in \left[3, 5\right]$ represents the level of noise. The quantitative results are show in Tab. \ref{tab:ablation} (left). Without inner body refining, the reconstruction performance drops fast as the noise level up. And by using our refining method, the performance improves and in general stable on different noise levels.

\noindent \textbf{Local part size}
We then study the effect of different radii for local part. To this end, we use our pretrained model, and test on the task of 4D reconstruction as previous. The comparisons are shown in Tab. \ref{tab:ablation} (right). As can be seen, the reconstruction performance is affected by the choice of part radius $r$. We choose $r=5cm$ in our experiment for slightly better results. We find that the over-small part is inclined to produce artifacts, possibly due to the limited receptive field within part. And the larger part could lead to overly smooth results.

\begin{table}

\centering
\caption{\label{tab:ablation}Ablation study. Left: the effectiveness of the inner body refining on different noise levels; Right: the effect of the part radius. We choose part radius $r=5cm$ in our experiments. The visualization examples are in Supp. Mat.}
\begin{minipage}{0.5\linewidth}
    \centering
	
    \renewcommand\arraystretch{0.8}
	\scalebox{0.7}{
	\begin{tabular}{p{1.5cm}<{\centering}p{2.5cm}<{\centering}p{1.5cm}<{\centering}p{1.5cm}<{\centering}p{1.5cm}<{\centering}}
	\toprule[1px]
	 Noise $\sigma$ & Refining & Ch.-$L_2$ $\downarrow$ & Normal $\downarrow$ & F-Score $\uparrow$ \\
     \midrule
    \multicolumn{1}{c|}{\multirow{2}{*}{3}} 
    & Before & 1.980 & 0.297 & 0.730  \\
    \multicolumn{1}{l|}{} & After & 0.628 & 0.245 & 0.776  \\
    \midrule
    \multicolumn{1}{c|}{\multirow{2}{*}{4}} 
    & Before & 5.469 & 0.382 & 0.605  \\
    \multicolumn{1}{l|}{} & After & 0.896 & 0.256 & 0.758  \\
    \midrule
    \multicolumn{1}{c|}{\multirow{2}{*}{5}} 
    & Before & 6.815 & 0.435 & 0.528  \\
    \multicolumn{1}{l|}{} & After & 0.753 & 0.260 & 0.733  \\
     \toprule[1px]
	\end{tabular}
	}
\end{minipage}
\hfill
\begin{minipage}{0.43\linewidth}

\renewcommand\tabcolsep{3pt}
\renewcommand\arraystretch{1.3}
    \scalebox{0.7}{
    \begin{tabular}{p{1.5cm}<{\centering}p{1.5cm}<{\centering}p{1.5cm}<{\centering}p{1.5cm}<{\centering}}
     \toprule[1px]
     Radius $\mathbf{r}$ & Ch.-$L_2$ $\downarrow$ & Normal $\downarrow$ & F-Score $\uparrow$ \\
     
     \midrule
    3$cm$ & 0.406 & 0.278 & 0.858  \\
    5$cm$ & \textbf{0.306} & \textbf{0.204} & \textbf{0.908} \\
    8$cm$ & 0.346 & 0.205 & 0.905 \\
    10$cm$ & 0.373 & 0.210 & 0.896 \\
     \toprule[1px]
    \end{tabular}
    }
\end{minipage}
\end{table}

\section{Conclusion}
This work introduces LoRD, a local 4D implicit representation for dynamic human, which aims to optimize a part-level temporal network for modeling detailed human surface deformation, e.g. clothing wrinkles. LoRD is learned on a very small set of training data (less than 100 sequences). Once trained, it can be used to fit different types of observed data including sparse point clouds, monocular depth images via auto-decoding. LoRD is capable to reconstruct high-fidelity 4D human and outperforms the state-of-the-art methods.

\subsubsection{Acknowledgements}
This work was supported by Shanghai Municipal Science and Technology Major Projects (No.2018SHZDZX01, and 2021SHZDZX0103).
The corresponding authors are Xiangyang Xue, Yanwei Fu and Yinda Zhang.

\clearpage
%
%
\bibliographystyle{splncs04}
\bibliography{egbib}

\clearpage

\begin{center}
\textbf{\Large Supplementary Material}
\end{center}

\setcounter{section}{0}

In this supplementary material, we provide implementation details, additional experimental results, visualization of the surface deformation within a local part, additional qualitative results, and discussions about limitations and future work of our approach. 

\section{Implementation Details}

\subsection{Network Architecture}

\begin{figure}[htb]
	\centering 
	\includegraphics[width=1\linewidth]{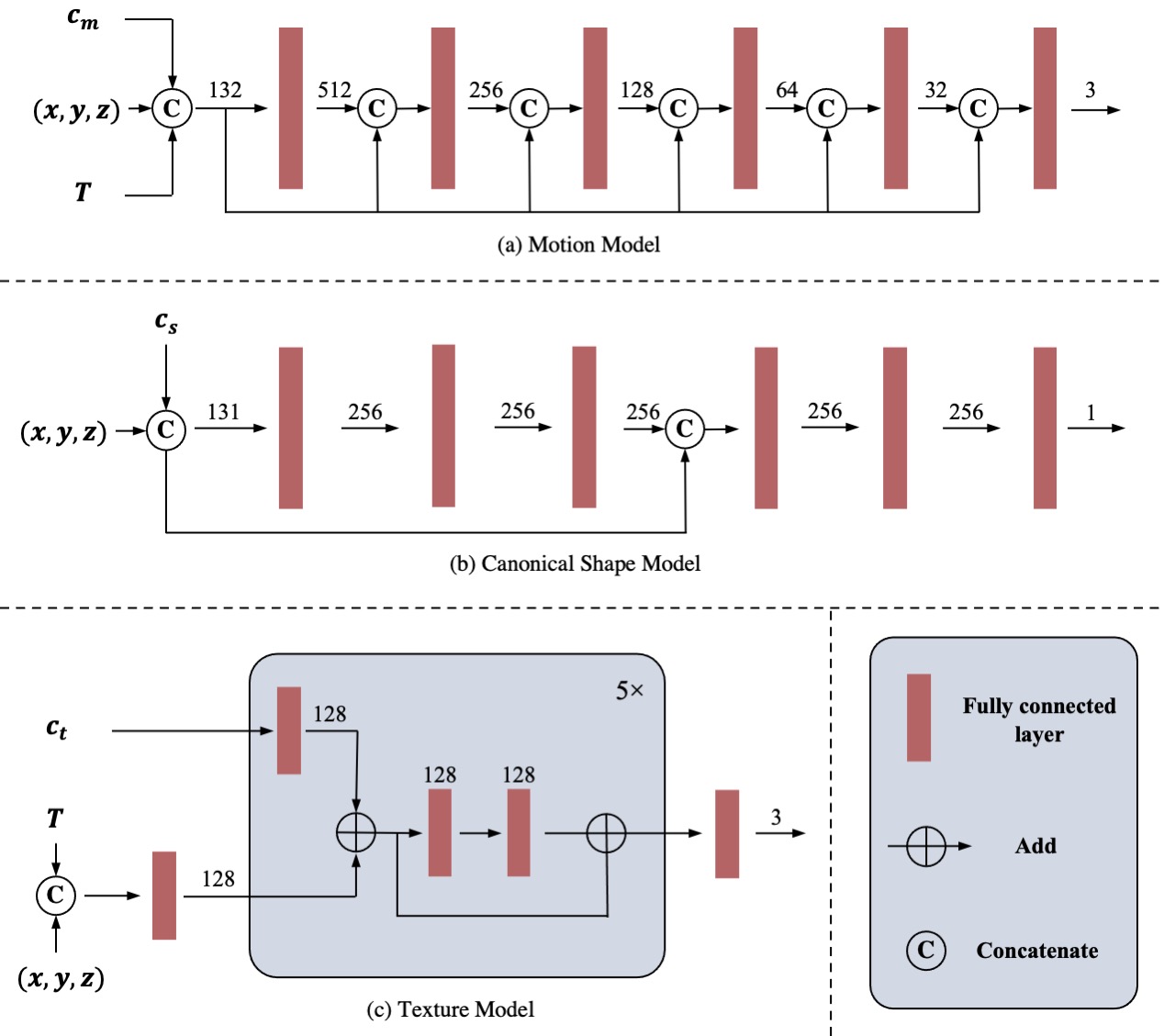}
	\caption{Detailed network architectures in our framework.}
	\label{fig:architecture}
\end{figure}

\noindent \textbf{Motion model}
We adapt the architecture of the IMNet \cite{chen2019learning} for our motion model. As shown in Fig. \ref{fig:architecture} (a), the input is the concatenation of: the motion code $c_m \in \mathbb{R}^{128}$, 3D query point $\mathbf{x} \in \mathbb{R}^{3}$ and normalized time value $T \in \mathbb{R}^{1}$. The network is based on multi-layer perceptrons with the skip connection to copy the input to concatenate with the output feature of the first 4 layers, each layer has the nonlinear activation of LeakyReLU ($\alpha=0.2$) \cite{maas2013rectifier} except the last layer. We follow LIG \cite{jiang2020local} to reduce the feature dimension of each hidden layer by 4 fold to obtain an efficient motion model. The output of our motion model is a deformation vector $\mathbf{x^{\ast}} \in \mathbb{R}^{3}$ that transforms the given point to its position in the space of the canonical frame, i.e. $T=0$.

\noindent \textbf{Canonical shape model}
The canonical shape model uses the auto-decoder network proposed in DeepSDF \cite{park2019deepsdf}, which is shown in Fig. \ref{fig:architecture} (b). The input is the concatenation of the canonical shape code $c_s \in \mathbb{R}^{128}$ and 3D query point $\mathbf{x} \in \mathbb{R}^{3}$, and the network predicts a signed distance value $\mathbf{s} \in \mathbb{R}^{1}$ for the given point. We reduce the number of hidden layers from 8 to 6 and the feature channels from 512 to 256 for the efficiency. And following IGR \cite{gropp2020implicit}, the softplus activation ($\beta=100$) and geometric initialization are also used.

\noindent \textbf{Texture model}
The texture model is used to produce the colored results in Fig. 1 and 5 of the main paper. We modify the decoder architecture proposed in Texture Fields \cite{oechsle2019texture} for our 4D texture inference, and the architecture is shown in Fig. \ref{fig:architecture} (c). The texture model is fed with the texture code $c_t \in \mathbb{R}^{128}$ and the concatenation of a 3D point $\mathbf{x} \in \mathbb{R}^{3}$ and a normalized time value $T \in \mathbb{R}^{1}$, and predicts the RGB values $\mathbf{c} \in \mathbb{R}^{3}$ for the given point.
There are five residual blocks in the texture model network, each of which consists of two fully connected layers with a skip connection from the input to the second layer. The input of each block is summed up with the feature encoded from the texture code $c_t \in \mathbb{R}^{128}$.

\subsection{Local Part Coordinates}
In this section, we introduce how to select part centers on SMPL meshes and define the local coordinate system for each part.

\noindent \textbf{Part center sampling}
Start from a template mesh of SMPL topology in the rest-pose, we first uniformly sample a point set $P$ ($|P|=100K$) on its surface, and then perform the following steps recurrently to maintain a set of the selected part centers $C$: 1) randomly choose a point $\mathbf{x}$ from $P$; 2) remove all the points whose Euclidean distance from $\mathbf{x}$ is less than or equal to $0.5\mathbf{r}$ from $P$, where $\mathbf{r}=5cm$ is the part radius we pre-defined; 3) add $\mathbf{x}$ to $C$. The loop terminates when $|P|=0$, and we finally obtain 2127 part centers.

\noindent \textbf{Local coordinate frame}
Given a point $\mathbf{c}_k$ on a triangle face as the part center, we use the face normal as the up-axis $\vec{a}_{k_1}$, the direction vector from point $c_k$ to a vertex of the triangle as another axis $\vec{a}_{k_2}$, and finally the last axis $\vec{a}_{k_3}=\vec{a}_{k_1} \times \vec{a}_{k_2}$. The rotation matrix of part $k$ is then define as $\mathbf{R}_{k}=\left[\vec{a}_{k_1}, \vec{a}_{k_2}, \vec{a}_{k_3}\right]$. Now, a 3D point $P_{glo}$ in the world coordinate frame can be transformed to $P_{loc}^k$ in the local coordinate frame of part $k$ with $P_{loc}^k=\mathbf{R}_{k}^T\left(P_{glo}-\mathbf{c}_k\right)$.

\subsection{Other Details}
\noindent \textbf{Point sampling}
The query points used for training and test-time optimization come from three sources: 1) surface; 2) near surface space; 3) free space in the bounding box. During training, we sample $M=10000$ surface points on the ground truth mesh, while at the test time, the points are randomly chosen from the input point cloud. Given the on-surface points, we obtain the near surface points by adding a displacement vector sampled from a Gaussian distribution $N\left(0,0.01\right)$ to each on-surface point. And $M/8$ free space points are uniformly sampled within the human bounding box. We compute the initial bounding box with the inner body mesh, and pad $10cm$ on each axis as the sampling region.

\noindent \textbf{Auto-decoding}
During the test time, we use the trained model to fit complete or partial point clouds via the auto-decoding manner \cite{park2019deepsdf} (main paper Sec. 3.4). Specifically, we fix the parameters of the local implicit network and optimize the latent codes with back-propagation by minimizing the objective function introduced in Sec. 3.3 of the main paper. 
We initialize the latent codes for each local part with the random vectors sampled from a Gaussian distribution $N(0,0.01)$ and use the Adam optimizer \cite{kingma2014adam} with learning rate $1e^{-3}$ to perform backward optimization for 3000 iterations. We use the same objective function and loss weights as training (Sec. 3.3 of the main paper). The optimization process takes around $15min$ for each sequence of $L=17$ frames on a single GeForce RTX 2080Ti GPU card.

\noindent \textbf{Mesh postprocessing}
As mentioned in the main paper (Sec. 3.4), the original extracted mesh of our method contains interior back-faces and some tiny floating components in the outside. This possibly because that for the off-surface points, the Eikonal term \cite{gropp2020implicit} only constrains the $L2$-norm of their gradients rather than specific SDF values, which may confuse the prediction of gradient direction on the points that far from the part centers. We remove these artifacts by using the post-processing algorithm introduced in LIG \cite{jiang2020local}. Specifically, we first compute the centroid and surface normal for each face of the original mesh reconstructed by the network, and find its $k$ nearest points on the input point cloud to 
calculate the mean normal consistency as the normal alignment score. Then a Laplacian kernel is used to smooth the normal alignment score, all faces with the score below a certain normal alignment threshold $n$ and disconnected components with area below $a$ are discarded. We used the same postprocessing parameters as LIG, except that we only preserve the most biggest connected component rather than use the area threshold $a$ as we focus on reconstructing single object.

\noindent \textbf{Mesh surface extraction}
Different from LIG \cite{jiang2020local}, which only evaluates the occupied grid cells and assumes all empty ones to be “exterior” space to extract the isosurface, we construct a SDF volume and evaluate every grid point with our local implicit function, ensuring to reconstruct the outer surface not covered by any local parts defined on the inner body mesh.
Similar to the strategy illustrated in the main paper Sec. 3.3, we evaluate a given point with $n$ parts that covered it ($n=32$ in the surface extraction process), and get the final prediction via average-pooling. And if not covered by any parts, $n$ nearest parts are used.
To make the inference more efficient, instead of cubic volume used in previous work \cite{mescheder2019occupancy,park2019deepsdf}, we resort to a rectangular volume that defined as the bounding box of inner body mesh with $10cm$ padding on each axis. It takes around $15s$ to extract a mesh of resolution 256.

\section{Additional Experimental Results}
\subsection{Non-Rigid Fusion}
\begin{figure}
	\centering 
	\includegraphics[width=1\linewidth]{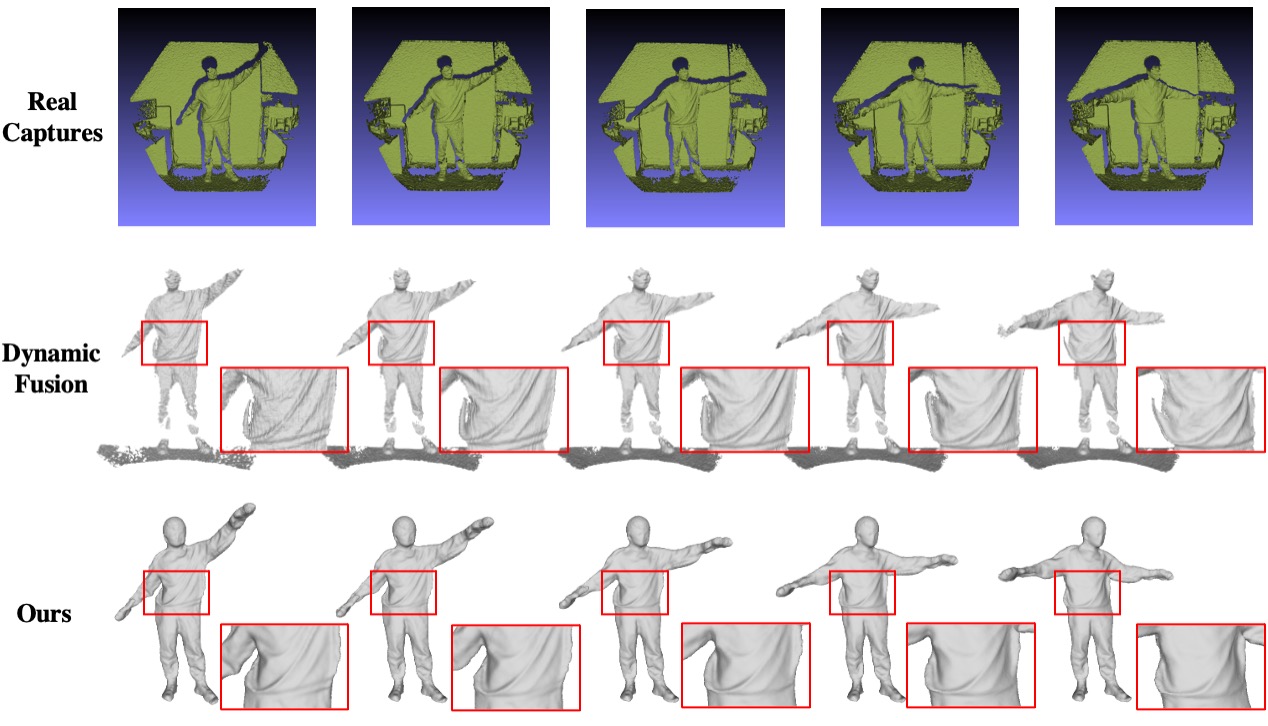}
	\caption{Monocular depth fusion on the real-world depth captures. The depth images are captured with a static Azure Kinect sensor. The first row shows the point cloud from raw depth captures. Our method is capable to recover the plausible geometry with detailed surface deformation, e.g. clothing wrinkles in the zoomed in part, in such challenging scenario.}
	\label{fig:real_data}
\end{figure}

\noindent \textbf{Real-world performance} To further demonstrate the value of our method in the actual application scenario, we perform extended evaluation on the real data. Specifically, we capture a human motion depth sequence with a static Azure Kinect sensor, and use our model pre-trained on 100 sequences to conduct non-rigid reconstruction based on the point cloud from raw depth images (note that the background is filtered out). The qualitative results are shown in Fig. \ref{fig:real_data}, DynamicFusion produces fine-grained surface details on cloth but contains noisy (body edge) and incomplete (face, arms and legs) areas. In contrast, our model recovers smoother and more complete geometry with plausible temporal deformation thanks to the local 4D formulation. We use part radius $r=10cm$ for more stable results.
Note that the experiments in here and Sec. 4.3 of the main paper, we only reconstruct the partial surface observed by the static camera in the time span, which cannot be quantitatively evaluated, thus we only show the qualitative comparisons.

\begin{figure}
	\centering 
	\includegraphics[width=0.95\linewidth]{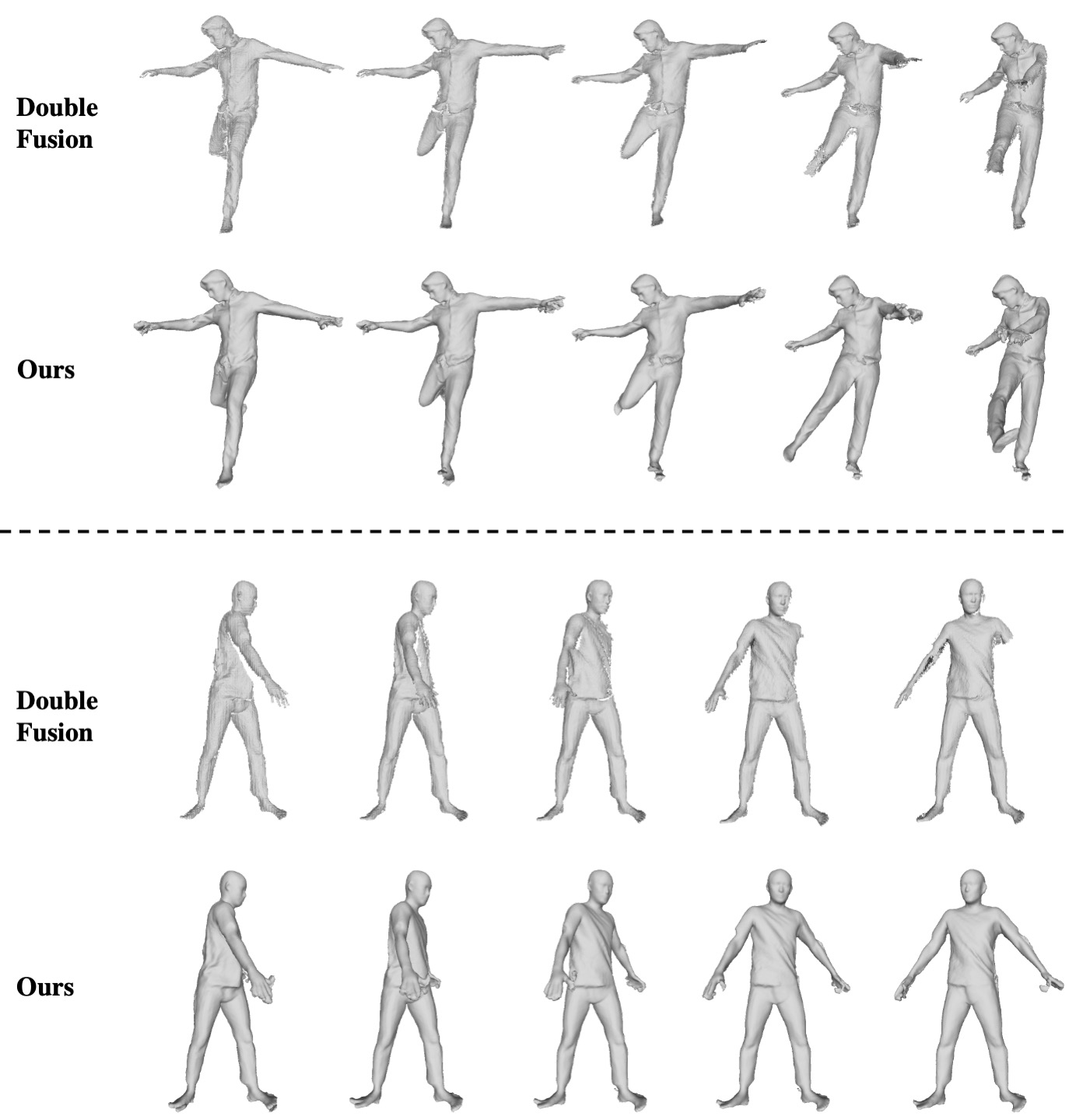}
	\caption{Qualitative comparisons with DoubleFusion. We choose two examples and uniformly sample 5 out of 17 frames for visualization.}
	\label{fig:doublefusion}
\end{figure}

\noindent \textbf{Comparisons with DoubleFusion}
Qualitative results are shown in Fig. \ref{fig:doublefusion}. We add the same SMPL body mesh used in LoRD into DoubleFusion as body prior, which generally makes the fusion reliable, but it still struggles with aligning finer motion. Also, there are some missing parts, e.g. arm or leg, caused by fitting error.
In contrast, LoRD employs a temporal model to provide global geometry consistency between frames, which is more robust to fitting error.

\noindent \textbf{Moving monocular camera}
In addition to the static camera, we also conduct non-rigid depth fusion in the setting of a moving monocular camera, that is, a person performing some motions during a time span with a depth camera rotating around him concurrently. In this case, the observation of each time step is still partial, but almost every body part can be observed during the time span. Our goal is to compensate each frame based on the geometry information observed in other frames. Specifically, given a dynamic mesh sequence of $L=17$ frames, we make the camera rotates around the up-axis (Y-axis in our setup) and render a $512\times512$ depth image every 360/17 degrees. In this experiment, we assume the camera poses are known, and use the same 10 novel sequences chosen in the generalizable 4D reconstruction task for evaluation.
We choose PTF \cite{wang2021locally}, NPMs \cite{palafox2021npms} and CAPE \cite{CAPE:CVPR:20} as our baseline. For PTF, we input the partial point cloud to the pretrained single-view model, and obtain the reconstructions with feed-forward fashoin. And for NPMs and CAPE, we use the auto-decoding manner to optimize the latent codes in their formulation based on each frame partial observation. Note that we provide the ground truth SMPL pose to CAPE, and only optimize the cloth latent code. All these methods are working in the framewise manner to produce the sequence results.
The quantitative results in Tab. \ref{tab:moving_mono} show that our LoRD representation outperforms all the baseline methods by a large margin. The qualitative comparisons are shown in Fig. \ref{fig:moving_mono}.
Thanks to the local 4D representation, the proposed method achieves high-quality completion results, and produces the detailed geometry and plausible temporal deformation in the invisible areas, while PTF and NPMs  fail to hallucinate accurate geometry from partial observations as they do not utilize the temporal information, and CAPE cannot model the high-fidelity surface details.

\begin{table}
\caption{\label{tab:moving_mono}Quantitative comparisons on monocular depth fusion with a moving camera. Our method outperforms all the baseline methods by a large margin.}
    \centering
    \renewcommand\tabcolsep{5pt}
    \begin{tabular}{lcccc}
     \toprule[1px]
      & Ch.-$L_2$ $\downarrow$ & Normal $\downarrow$ & F-Score $\uparrow$ \\
     \midrule
    PTF \cite{wang2021locally} & 37.936 & 0.810 & 0.145  \\
    NPMs \cite{palafox2021npms} & 0.981 & 1.933 & 0.429  \\
    CAPE \cite{CAPE:CVPR:20} & 1.010 & 0.356 & 0.377 \\
    Ours & \textbf{0.395} & \textbf{0.226} & \textbf{0.750}  \\
     \toprule[1px]
    \end{tabular}
\end{table}

\subsection{Interpolation of Latent Codes}
\begin{figure}
	\centering 
	\includegraphics[width=1\linewidth]{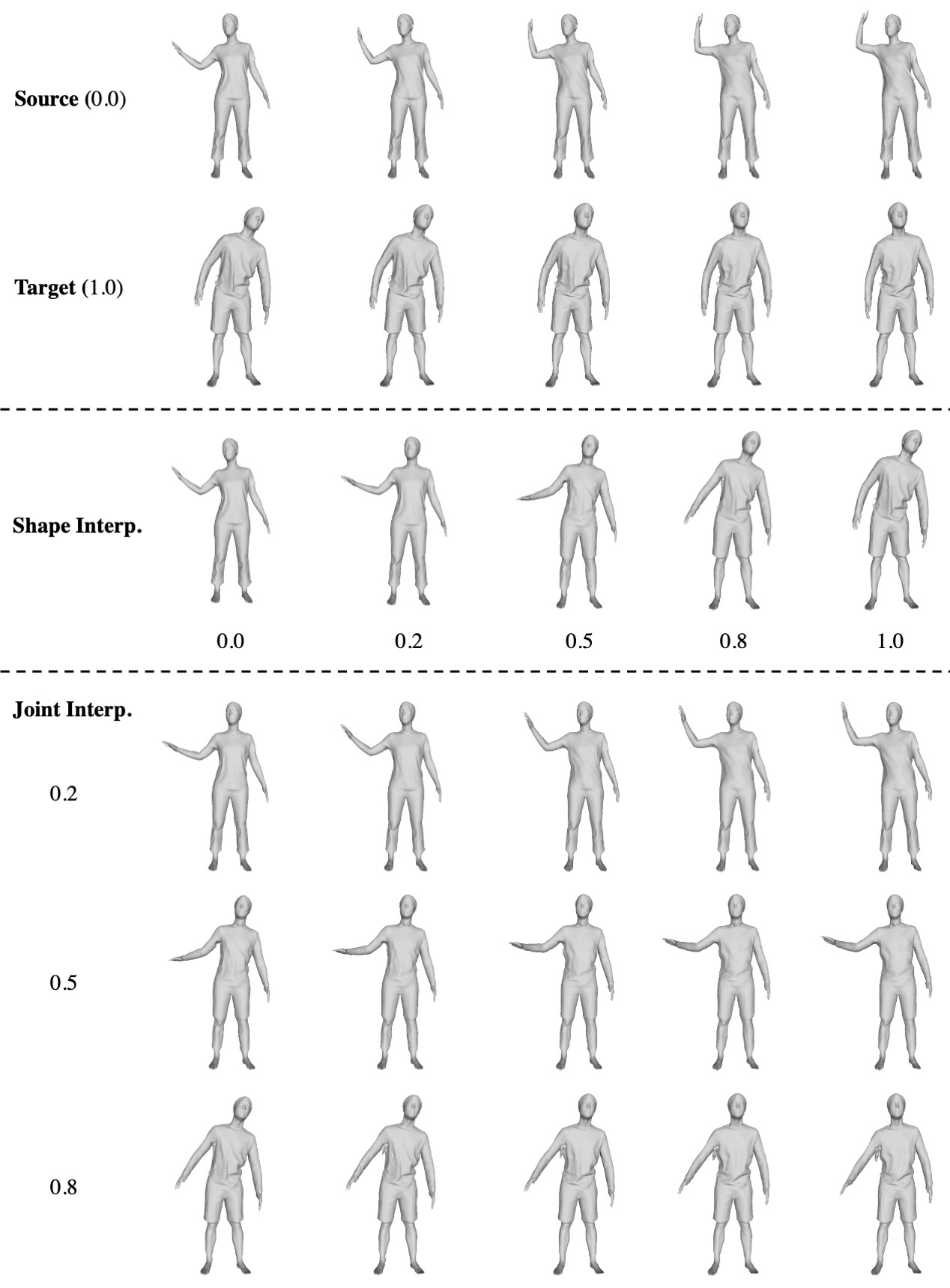}
	\caption{We choose two sequences (Source and Target) to perform interpolation of latent codes. ``Shape Interp.'' means we interpolate latent shape codes and SMPL poses of the first frames to show the evolution of shape. ``Joint Interp.'' indicates we interpolate shape code, motion code and per frame pose jointly. The decimal values denote the interpolation coefficients}
	\label{fig:latent_interpolation}
\end{figure}

Like many other local implicit representations, our model does not have a compact global latent space to sample from, so the representation ability is often measured by the capability of fitting observations, validated in Sec 4.1 of main paper.
Nevertheless, we can still interpolate between two human sequences by interpolating representation between the corresponding local parts (as shown in Fig. \ref{fig:latent_interpolation}). We first linearly interpolating $c_s$ and SMPL poses showing smooth change between two subjects. Then we interpolate $c_m$, $c_s$ and per frame SMPL pose jointly to show the representation ability of motion space, since the local deformation is related to global motion. Note that the texture model is optimized per sequence without continuous latent space.

\subsection{Ablation Study}
\noindent \textbf{Effect of sequence length $L$} 
The key novel capability of our model is to represent the temporal deformation of 3D shape, so that setting $L=1$ makes this infeasible, and our model degenerates to framewise method LIG \cite{jiang2020local} and thus achieves similar performance. Additionally, we test different $L$ on instance-level reconstruction from sparse points task and show the results in Tab. \ref{tab:seq_length}. Note that longer sequence demands stronger network capacity while more geometry information can be exchanged between frames through our motion model. The results show that LoRD is able to work with different sequence lengths. We choose $L=17$ following 4D-CR \cite{jiang2021learning} during training to learn our 4D representation. For longer sequence during test-time, we can utilize the sliding window strategy similar to HMMR \cite{kanazawa2019learning} to recurrently recover the whole sequence.

\begin{table}
    \caption{\label{tab:seq_length} Evaluations of LoRD on different sequence lengths.}
        \centering
        \renewcommand\tabcolsep{5pt}
    \begin{tabular}{lcccc}
    \toprule[1px]
    $L$ & Ch.-$L2$ $\downarrow$ & Normal $\downarrow$ & F-Score $\uparrow$ \\
    \midrule
    5 & 0.175 & 0.160 & 0.914 \\
    10 &  0.207 & 0.176 & 0.880 \\
    17 & 0.192 & 0.158 & 0.945 \\
    20 & 0.159 & 0.173 & 0.875 \\
    30 & 0.389 & 0.221 & 0.718 \\
    \toprule[1px]				
    \end{tabular}
\end{table}

\noindent \textbf{Generalization} 
Besides the results shown in Sec. 4.2 of the main paper, we also choose 10 sequences from our training set, and get the performance (Ch.-$L_2$=0.317, Normal=0.170, F-Score=0.926) on 4D reconstruction task. 
The results are in general comparable with the performance on testing set (last row of Fig. 6 (a, II) in the main paper), which reflects that thanks to the local part formulation, our model has strong generalization ability with the prior of local surface deformation learned from the training sequences. We further verify this by training our model on even 1 motion sequence of 17 frames, which also can produce the high-quality reconstructions on novel sequences. We show the qualitative results in Sec. 4.2 of Supp. Mat.

\begin{figure}
	\centering 
	\includegraphics[width=1\linewidth]{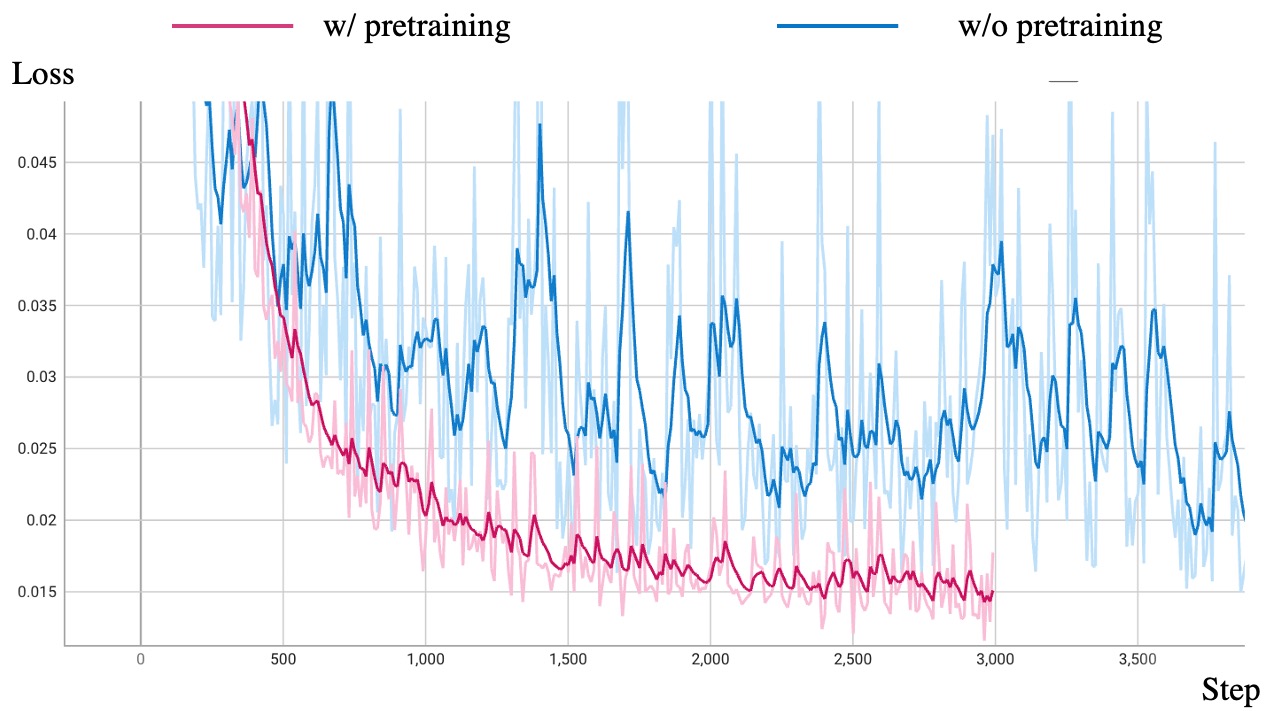}
	\caption{Loss curves during optimization w/ and w/o pretraining.}
	\label{fig:loss_curve}
\end{figure}

\noindent \textbf{Comparison with only test-time training}
In Fig. \ref{fig:loss_curve}, we show the loss curves of optimization w/ and w/o pretraining. It can be seen that the pretraining helps optimization converge faster and more stable. By taking about 50 mins per sequence, optimizing from scratch obtains slightly better performance than optimizing with pretraining.

\section{Local Surface Visualization}
\begin{figure}[htb]
	\centering 
	\includegraphics[width=1\linewidth]{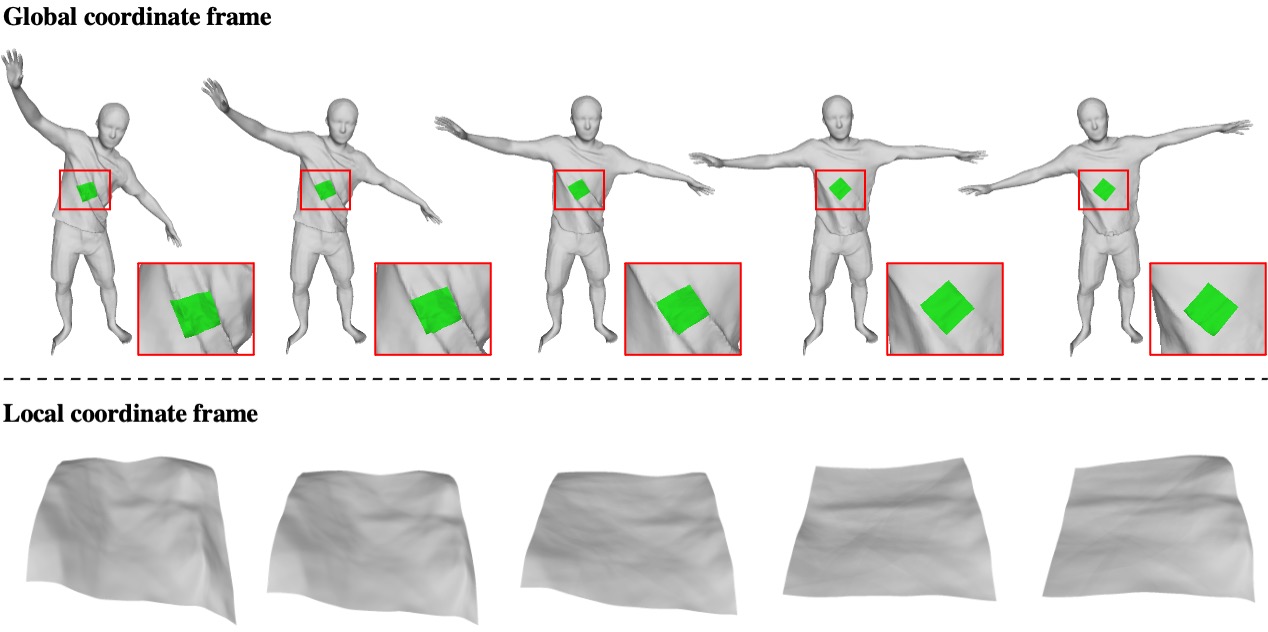}
	\caption{We select a local part and visualize the surface patch within it. The results above show the temporal deformation of this patch (green) under the global coordinate frame, while below under the local coordinate frame.}
	\label{fig:vis_patch}
\end{figure}

Our LoRD representation aims to use a local part-level network to model the detailed temporal deformation of surface patches. To show this, we visualize the temporal deformation of a local patch in Fig. \ref{fig:vis_patch}. Specifically, we choose a point cloud sequence of 17 frames, each of which has $10K$ points, and perform auto-decoding to optimize the latent codes of each local part. After that, we select a part and extract the surface patch with the local implicit network (main paper Sec. 3.2) conditioned on its latent codes. We show the temporal deformation of this local patch under the global (above) and local (below) coordinate frames respectively. It can be seen that the temporal changing within the local part is smooth and coherent, and our method successfully models the detailed deformation, e.g. changing of the clothing wrinkle.

\section{Qualitative Results}
\subsection{4D Reconstruction}
\noindent \textbf{Shape quality}
Fig. \ref{fig:4d_instance}, \ref{fig:4d_instance2} and \ref{fig:4d_generalize} are the extended figures of Fig. 5 in the main paper, which show more qualitative comparisons with the SoTA methods on 4D reconstruction from sparse points. And Fig. \ref{fig:4d_generalize_more} shows some additional reconstruction results of our model trained on 100 sequences.

\begin{figure}[htb]
	\centering 
	\includegraphics[width=1\linewidth]{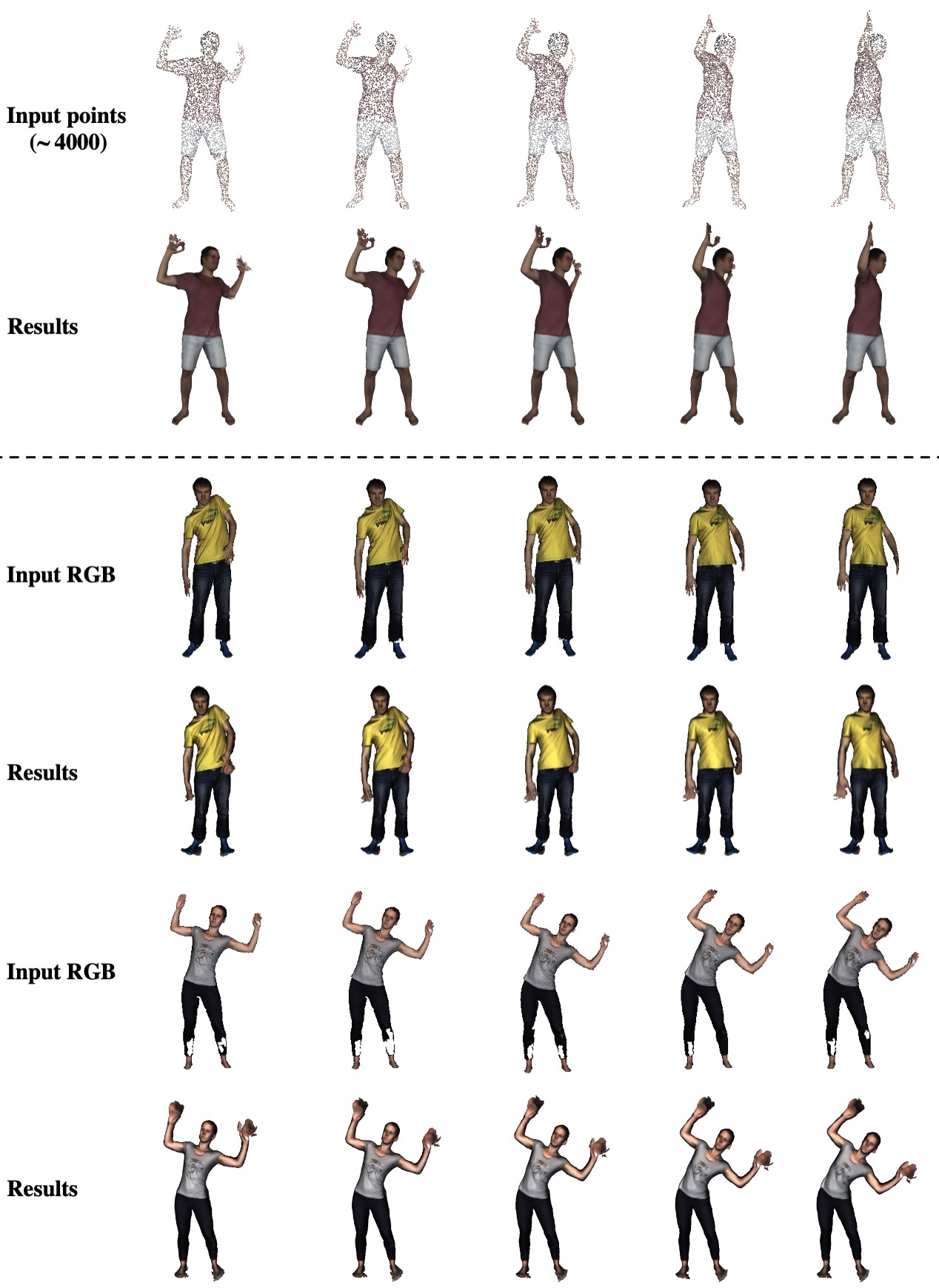}
	\caption{More textured results achieved by our method. Note that the results below are obtained from RGB-D inputs, we only show color images for visualization.}
	\label{fig:texture}
\end{figure}

\begin{figure}
	\centering 
	\includegraphics[width=1\linewidth]{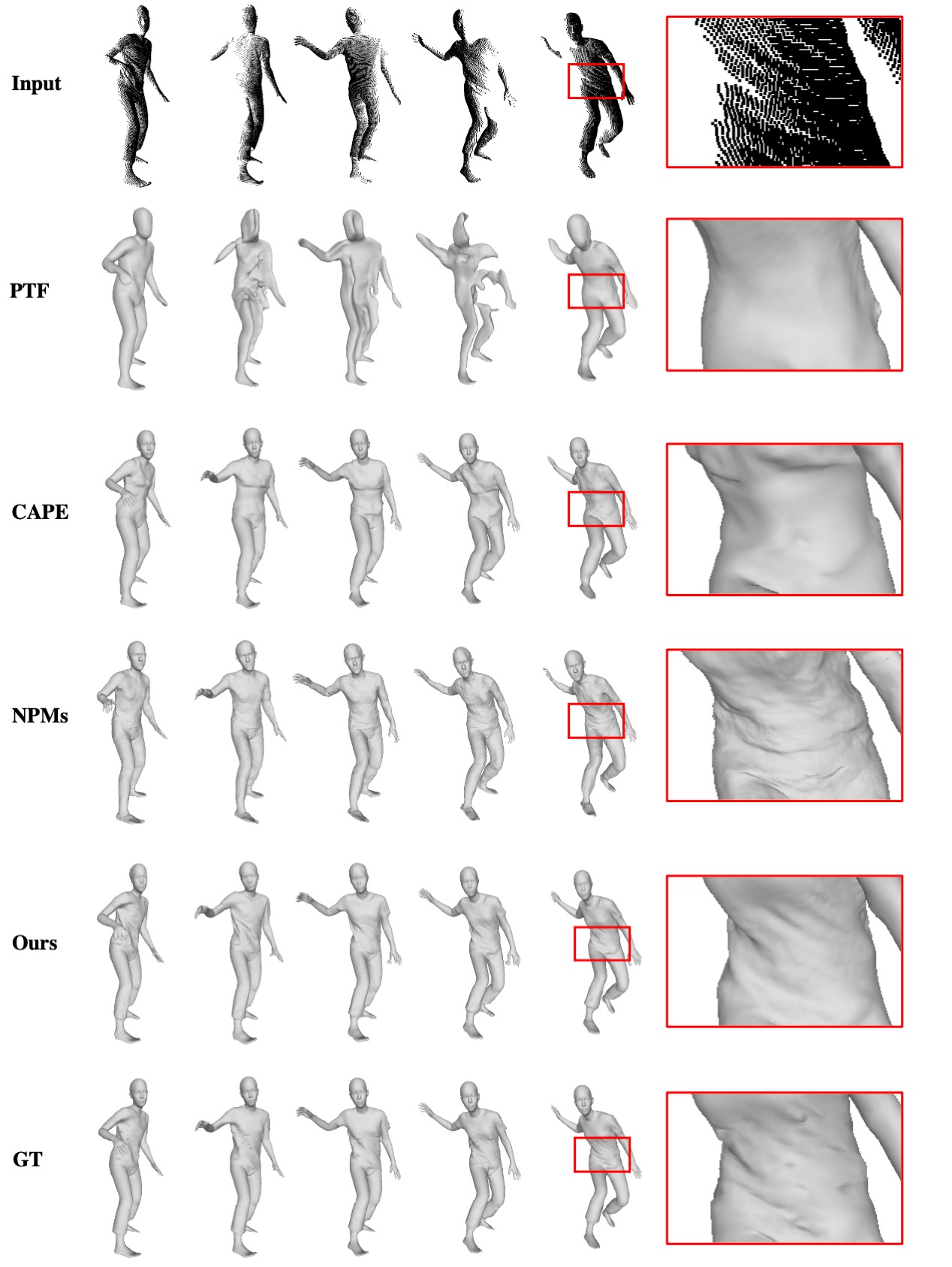}
	\caption{Qualitative results on monocular depth fusion with a moving camera. The reconstructed sequences have $L=17$ frames, and we uniformly choose 5 frame for visualization. We assume a moving camera that rotates around the performer, and render a depth image every 360/17 degrees. The partial point clouds are obtained by back-projecting the depth images with the camera intrinsics, and rotate according to the known camera extrinsics.}
	\label{fig:moving_mono}
\end{figure}

\begin{figure}[tb]
	\centering 
	\includegraphics[width=0.95\linewidth]{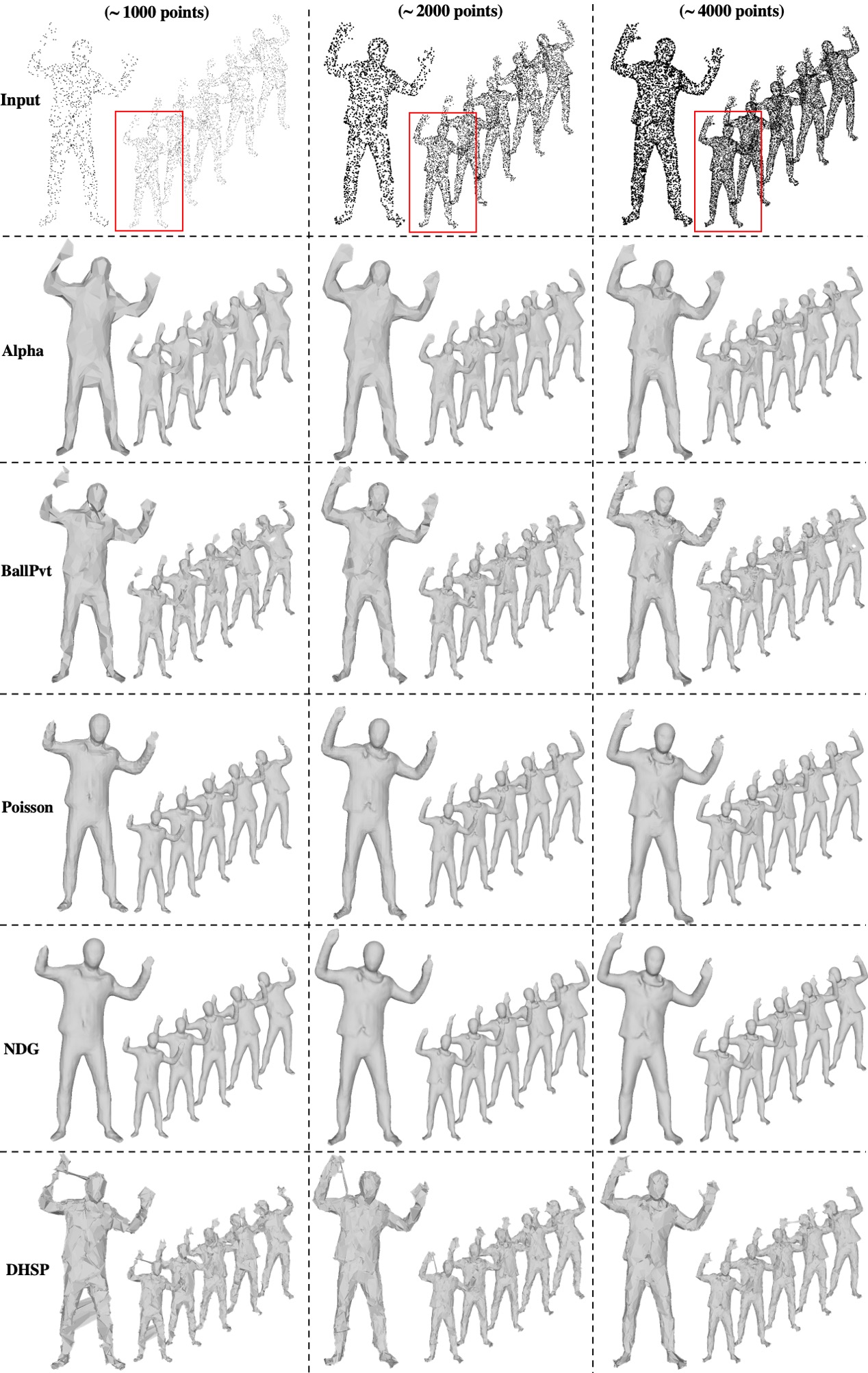}
	\caption{4D reconstruction from sparse points (instance-level) (1).}
	\label{fig:4d_instance}
\end{figure}

\begin{figure}[htb]
	\centering 
	\includegraphics[width=0.95\linewidth]{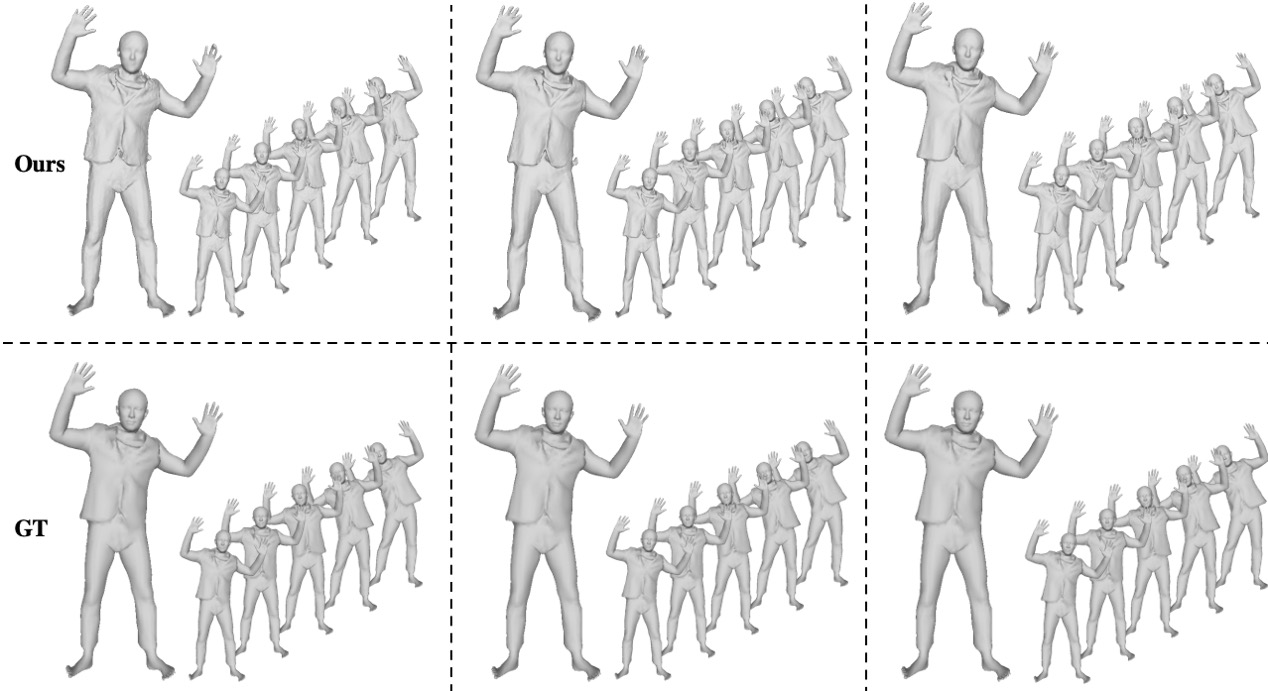}
	\caption{4D reconstruction from sparse points (instance-level) (2). The reconstructed sequences have $L=17$ frames, and we uniformly choose 5 frame for visualization. We zoom in the first frame to show the surface details clearer.}
	\label{fig:4d_instance2}
\end{figure}

\noindent \textbf{Textured results} 
We show more textured results in Fig. \ref{fig:texture}. We choose raw scan mesh sequences from CAPE dataset (containing holes and noises) and use our pretrained LoRD model in this experiment. The results above are obtained with colored sparse point clouds sampled from the scan meshes as input. And the results below are obtained from the rendered RGB-D sequences.

\subsection{Ablation study} \label{sec:supp_ablation}
\begin{figure}
	\centering 
	\includegraphics[width=0.9\linewidth]{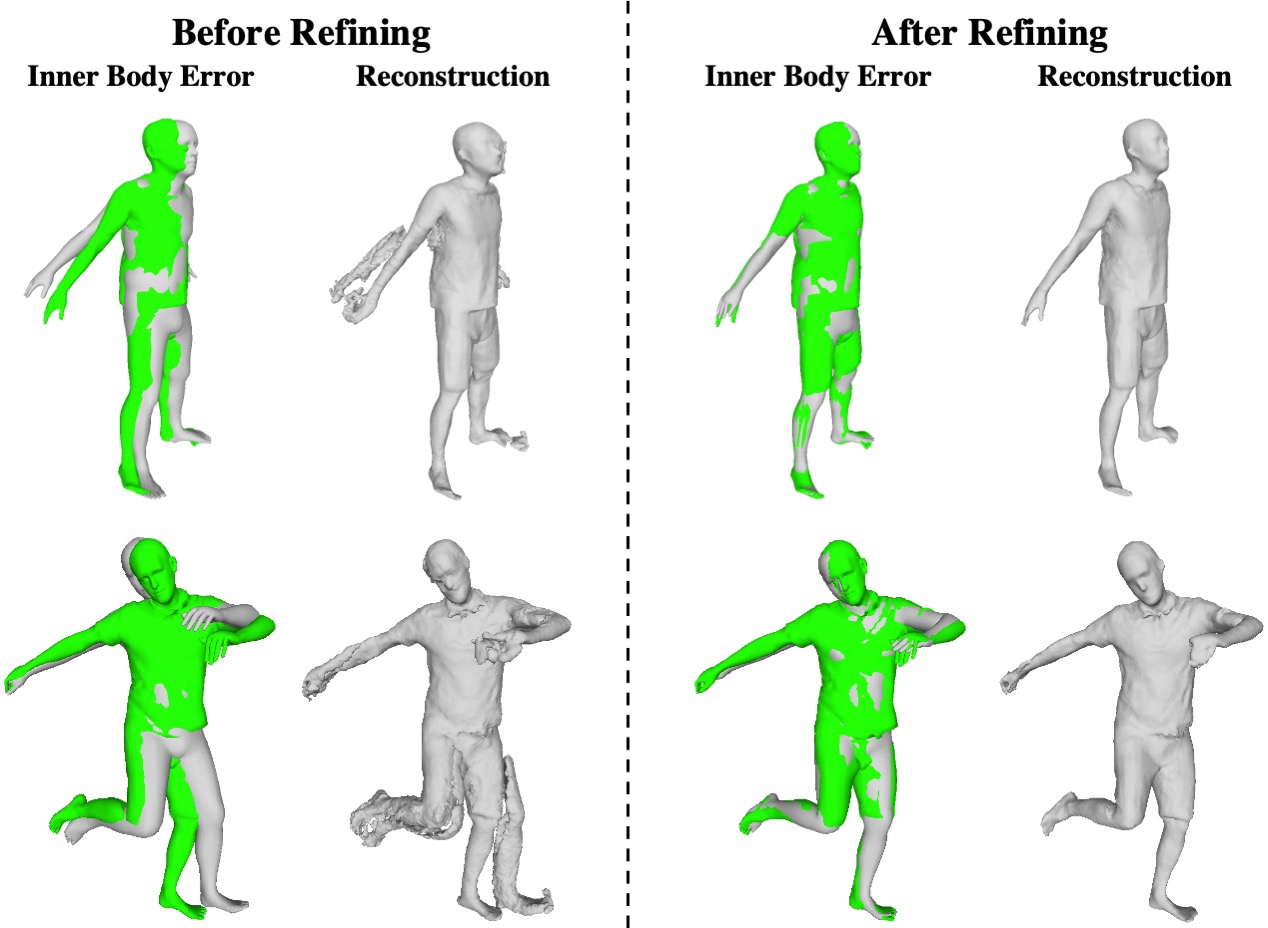}
	\caption{Effectiveness of the inner body refining. We show the inner body and the reconstructed meshes before (left) and after (right) our inner body refining process. Note that we stack the ground truth clothed mesh (green) with the inner body estimation (gray) to show the inaccurate parts.}
	\label{fig:refining}
\end{figure}

\noindent \textbf{Imperfect body tracking}
Fig. \ref{fig:refining} (before refining) shows some challenging cases that LoRD produces artifacts with inaccurate body tracking. We provide the reconstructions after refining that demonstrate the effectiveness of our inner body refining method. For each example, we stack the ground truth clothed mesh (green) together with the initialized inner body mesh (gray) on the left to reflect the inaccurate estimation, and show the reconstructions on the right. The results show our refining process successfully corrects the noisy inner body, which facilitates the reconstruction. 

\noindent \textbf{Local part size}
Fig. \ref{fig:part_size} shows the reconstruction results of different part radii. We can observe that the part size slightly affects the quality of reconstructions, the over-small ($\mathbf{r}=3cm$) part produces some artifacts around body and hands and the reconstructed surface is under-smooth, whereas the larger parts ($\mathbf{r}=8cm$\&$\mathbf{r}=10cm$) tend to recover overly smooth results for some frequency details such as clothing wrinkles and fingers.

\begin{figure}
	\centering 
	\includegraphics[width=0.95\linewidth]{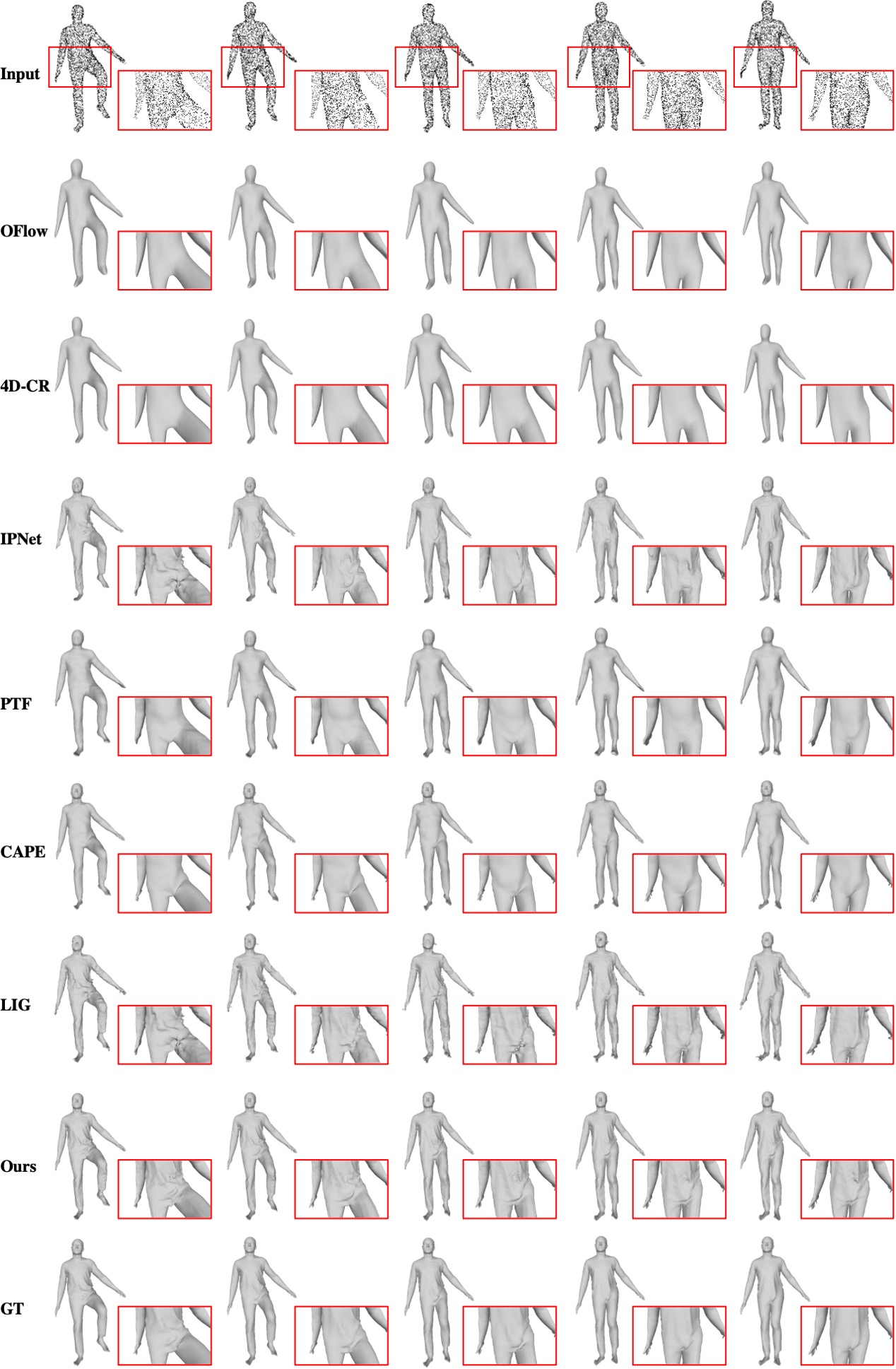}
	\caption{4D reconstruction from sparse points (generalization). The reconstructed sequences have $L=17$ frames, and we uniformly choose 5 frame for visualization.}
	\label{fig:4d_generalize}
\end{figure}

\begin{figure}
	\centering 
	\includegraphics[width=0.9\linewidth]{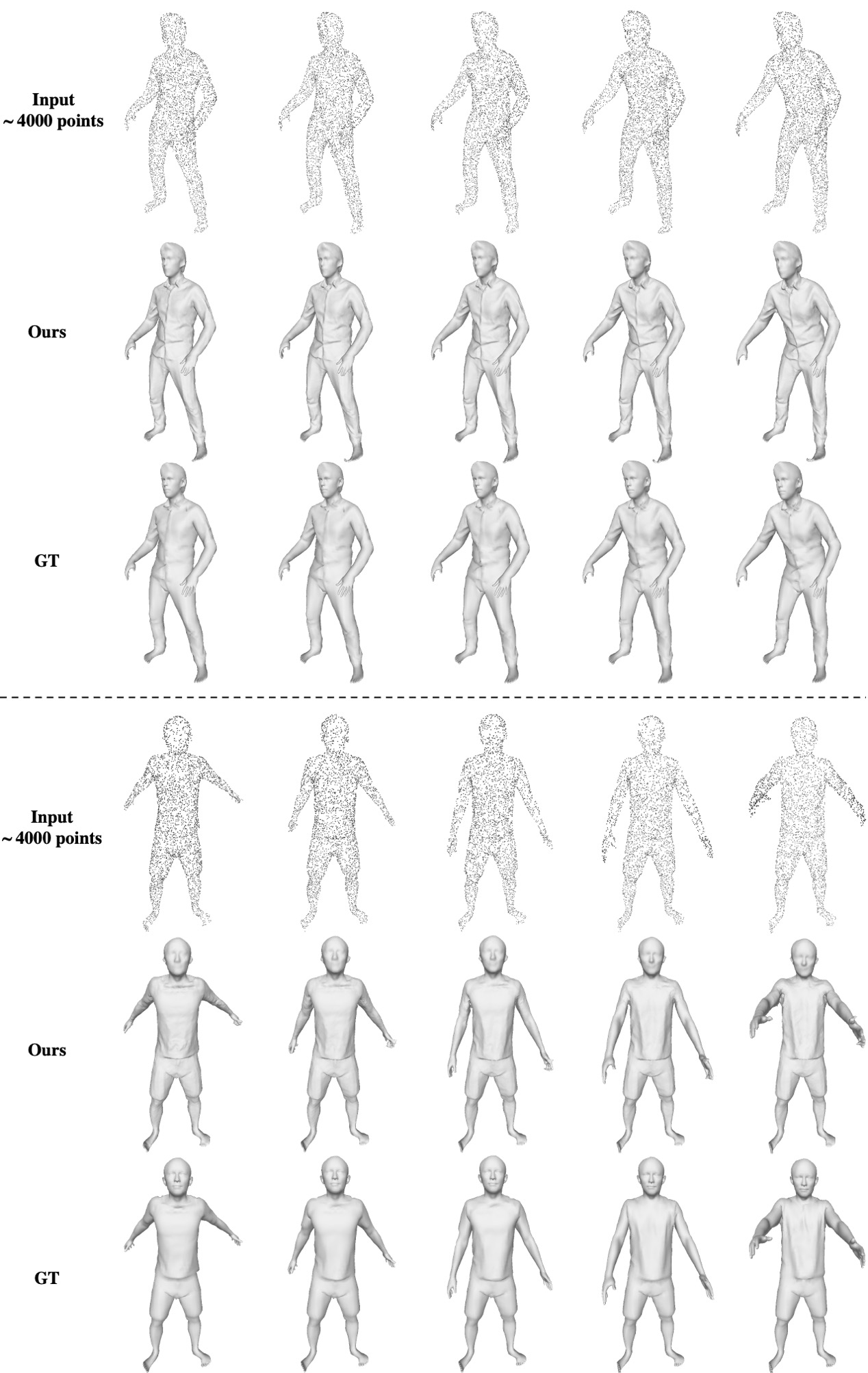}
	\caption{4D reconstruction from sparse points (generalization). Here we show more qualitative results achieved by our model trained on 100 sequences. The reconstructed sequences have $L=17$ frames, and we uniformly choose 5 frame for visualization.}
	\label{fig:4d_generalize_more}
\end{figure}

\begin{figure}[tb]
	\centering
	\includegraphics[width=1\linewidth]{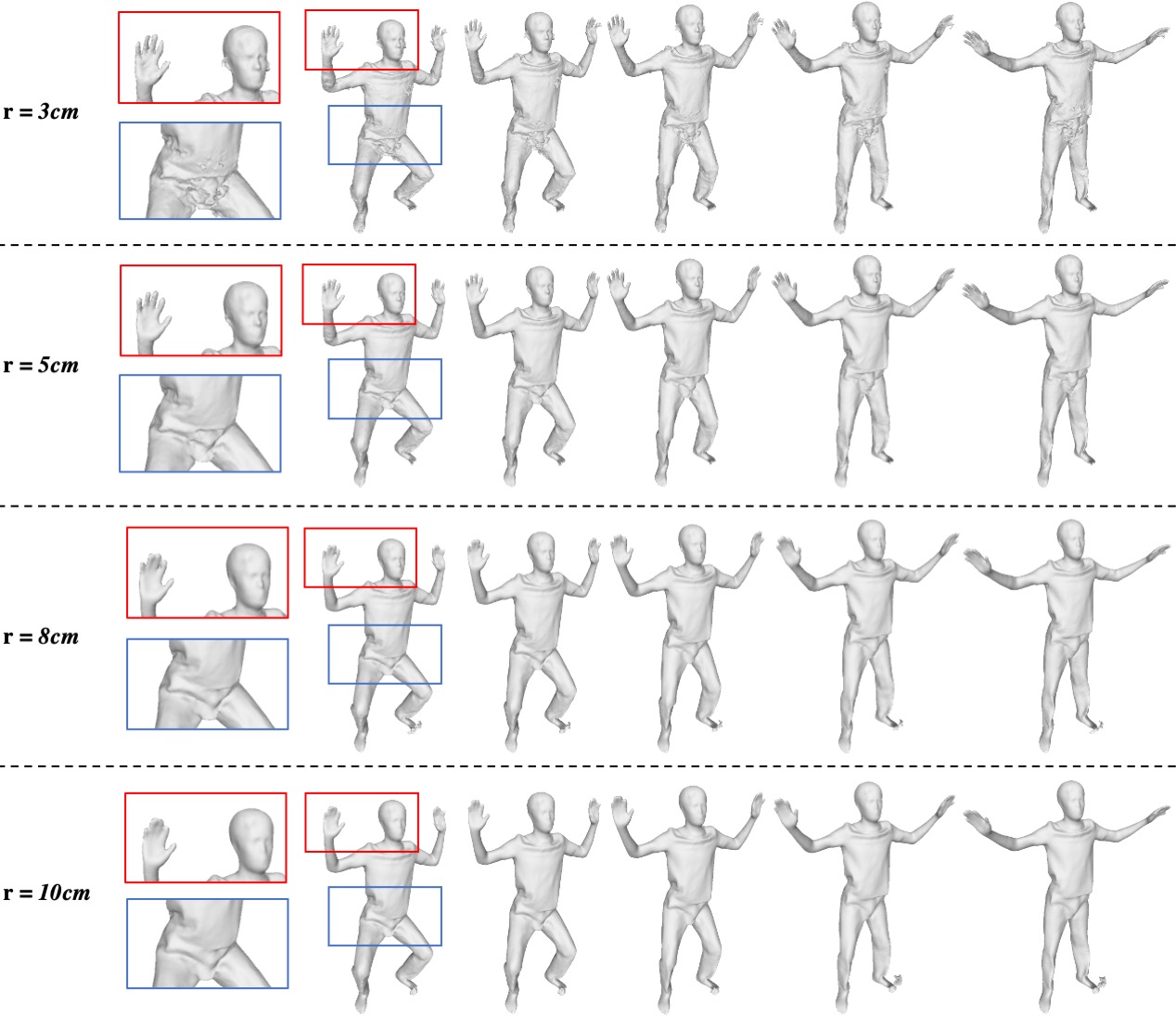}
	\caption{Effect of the part radius. Different sizes of local parts affect the reconstruction performance but not very heavily. We choose part radius $\mathbf{r}=5cm$ in our experiments as it in general produces better results.}
	\label{fig:part_size}
\end{figure}

\begin{figure}[tb]
	\centering 
	\includegraphics[width=0.9\linewidth]{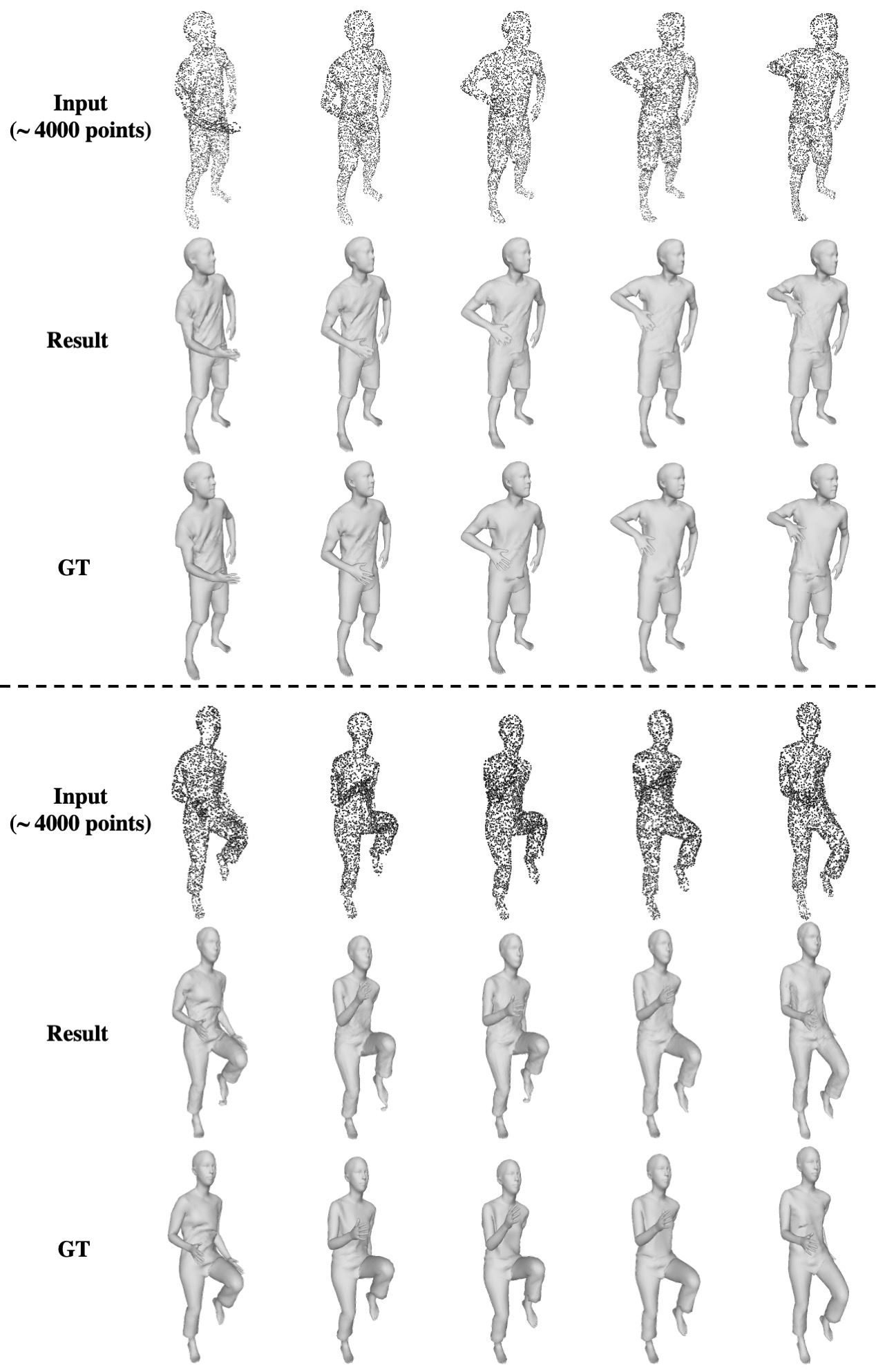}
	\caption{Qualitative results from our model trained on only 1 motion sequence of length $L=17$, which show that our LoRD representation can learn local deformation prior from very few data and generalize to novel sequences with high-quality geometry and temporal deformation.}
	\label{fig:ablation_gene}
\end{figure}

\noindent \textbf{Generalization}
As mentioned in the main paper Sec. 3.3, the training of our LoRD representation is very data-efficient. We show the results of our model trained on 100 sequences (Fig. 5, 6 (b) in the main paper and Fig. \ref{fig:4d_generalize}, \ref{fig:4d_generalize_more} in the Supp. Mat). Additionally, we train our model on one motion sequence of length $L=17$ from the training set, then test on the novel sequences, and show the qualitative results in Fig. \ref{fig:ablation_gene}. As shown, our model trained on even one sequence still gains generalization ability and produces the high-fidelity reconstructions, which demonstrates the generalization power of our local 4D representation.

\section{Limitations and Future work}
The proposed LoRD representation shows the powerful capability and achieves the state-of-the-art performance on various tasks. Now we discuss a few limitations of our method which also points to the future directions. 

First, we now rely on the SMPL body model to temporally track local parts, though existing methods can produce accurate body in many cases, it is still challenging to work in the complex real scenario. Extending our representation to cooperate with more general tracking methods such as scene flow, deformation graph, would make our method stronger and capable of modeling non-human objects, e.g. animals.

Second, the current experiments mainly focus on 4D reconstruction from 2.5/3D data, e.g. sparse point clouds, RGB-D. Combining the recent neural rendering techniques \cite{mildenhall2020nerf,wang2021neus,yariv2021volume} with our LoRD representation to support 4D reconstruction from pure RGB videos would be a promising future direction.

Third, we currently use a unified part radius in our formulation. However, different body parts contain various levels of detail, which may be suitable to model by different sizes of parts. Defining the part radii according to the body part label would be one solution.

We believe that the proposed representation could potentially be a building block for various applications, e.g. Metaverse, Robotics, animation, and provides some insights for future research directions.

\end{document}